\documentclass[preprint,3p]{elsarticle}

\usepackage{amsmath}
\usepackage{amssymb}
\usepackage{amsfonts}
\usepackage{graphicx}
\usepackage{textcomp}
\usepackage{xcolor}
\usepackage{booktabs}
\usepackage{tabularx}
\usepackage{longtable}
\usepackage{multirow}
\usepackage{array}
\usepackage{rotating}
\usepackage{adjustbox}
\usepackage{threeparttable}
\usepackage[referable]{threeparttablex}
\usepackage{algorithm}
\usepackage{algpseudocode}
\usepackage[hidelinks]{hyperref}
\usepackage{xurl}
\biboptions{sort&compress}
\newcommand{\nameModel}{AgroVisNet}
\newcommand{\nameDataset}{BD-PlantDX}
\newif\ifshowtodo
\showtodofalse
\journal{Computers and Electronics in Agriculture}
\begin{document}
\begin{frontmatter}
\title{\nameModel: A lightweight Convolutional Network and the \nameDataset{} Expert-Validated Benchmark for Radish, Potato and Pointed Gourd Disease Classification}
\author[baust]{Md. Abdullah Mandal\corref{cor1}}
\ead{abdullahmandalm@gmail.com}
\author[ruet]{Saad Ahmed}

\cortext[cor1]{Corresponding author.}

\author[baust]{Md. Khalid Syfullah}

\affiliation[ruet]{
    organization={Department of Computer Science and Engineering, Rajshahi University of Engineering \& Technology},
    city={Rajshahi},
    postcode={6204},
    country={Bangladesh}
}

\affiliation[baust]{
    organization={Department of Computer Science and Engineering, Bangladesh Army University of Science \& Technology},
    city={Saidpur},
    country={Bangladesh}
}
\begin{abstract}
Automated plant disease diagnosis is increasingly deployed on farmer-held devices in regions where agronomic expertise is scarce and network connectivity is unreliable. Three obstacles limit its practical value: public benchmarks are dominated by a small set of non-native crops, region-specific datasets are rarely validated by domain experts, and the architectures that reach competitive accuracy carry parameter budgets that are unsuited to low-cost hardware. We propose \nameModel, a compact convolutional network trained from scratch, together with \nameDataset, an expert-validated benchmark of 12,432 field images spanning 12 classes of radish, potato and pointed gourd in healthy and diseased states, collected across the Bogura and Nilphamari districts of Bangladesh. \nameModel{} couples grouped bottleneck residual blocks carrying sequential channel and spatial attention with multi-scale depthwise blocks and a dual-pooling classification head, reaching 290,572 trainable parameters. On \nameDataset{} the model attains 99.52\% test accuracy and 99.52\% weighted F1, exceeding all six ImageNet-pretrained lightweight backbones evaluated under an identical protocol while using 8.7 to 16.8 times fewer parameters and 1.3 to 8.5 times fewer multiply--accumulate operations. Exported for deployment, the model quantises to a 0.46 MB full-integer network at a 0.22 percentage-point accuracy cost and classifies an image in 8.40 ms on a single CPU. Across five random seeds accuracy remains at $99.57 \pm 0.10$\%, a ten-variant ablation isolates the contribution of each component, and the same architecture transfers without redesign to two independently collected datasets at 98.71\% and 99.05\% accuracy. Grad-CAM evidence indicates that predictions rest on lesion-bearing leaf regions rather than on background cues.

\end{abstract}
\end{frontmatter}

\section{Introduction}
\label{sec:introduction}
Timely identification of crop disease governs both yield and the quantity of chemical input a grower applies. Radish (Raphanus sativus) is a commonly grown vegetable, especially in Asian nations, and holds significant importance in human diet, local farming, and economic activities~\citep{nishio2017economic}. Late detection of potato  disease in Bangladesh was associated with average yield losses of 25 to 57\% over the 2016 to 2019 seasons, placing these pathologies among the most economically damaging in the country~\citep{islam2022identification}. Manual scouting, the prevailing practice, depends on visual inspection by trained personnel and delivers neither the coverage nor the response time that early intervention requires~\citep{sladojevic2016deep}. Domain specialists able to differentiate fungal lesions, insect damage and nutrient disorders are scarce and expensive to reach from remote holdings, which leaves growers to apply broad-spectrum pesticide on the basis of an uncertain diagnosis~\citep{qadri2024advances}.
Computer vision offers a route around this constraint. Convolutional neural networks trained on leaf imagery now report accuracies above 99\% on public benchmarks, and a smartphone camera is sufficient to acquire the input~\citep{geetharamani2019identification, barman2020comparison}. Molecular and spectroscopic alternatives, including DNA-based point-of-care assays~\citep{lau2017advanced}, mass spectrometry imaging~\citep{ajith2022mass} and near-infrared spectroscopy~\citep{tan2021ganoderma}, achieve high specificity, yet their cost and procedural complexity keep them outside the reach of a smallholder. Image-based diagnosis therefore remains the only modality that combines diagnostic value with a deployment cost a grower can absorb.
Two gaps persist across the published work in this area. The first concerns data. A large share of reported results is obtained on PlantVillage~\citep{hassan2021identification, rashid2021multi, rozaqi2020identification}, a laboratory-acquired collection of 54,305 images covering 14 crop species, none of which is native to the cropping systems of northern Bangladesh. Models fitted to that distribution encounter a different pathogen profile, a different cultivar set and a different imaging environment once deployed locally, and the few region-specific collections that exist are typically limited to one crop and carry no record of agronomist verification~\citep{hasan2025smartphone, banerjee2023integrated}. The second gap concerns capacity. Reported accuracy gains are frequently obtained by fine-tuning backbones of 5 to 25 million parameters~\citep{pandian2022plant, reis2024potato}, a budget that resists deployment on the low-cost handsets that dominate rural Bangladesh, and the compact backbones proposed as an alternative are pretrained on ImageNet rather than designed for the fine-grained, texture-driven discrimination that lesion classification demands.
Existing work addresses at most one of these gaps at a time. Region-specific datasets are released without a matching efficient architecture~\citep{hasan2024comprehensive}, and compact architectures are validated only on laboratory imagery of non-native crops~\citep{vishnoi2022detection}. To the best of our knowledge, no prior study jointly delivers an expert-validated multi-crop dataset for this cropping system and a sub-megabyte model that matches ImageNet-pretrained backbones on it. This paper closes both gaps together.
The contributions of this paper are fivefold:
\begin{itemize}
\item We introduce \nameDataset, an expert-validated benchmark of 12,432 images spanning 12 classes drawn from radish, potato and pointed gourd, each crop represented in healthy and diseased states. Acquisition covered the Bogura and Nilphamari districts of Bangladesh, class assignment was verified by practicing agronomists, and the collection includes root as well as foliar presentations.
\item We propose \nameModel, a convolutional network trained from scratch that combines grouped bottleneck residual blocks carrying sequential channel and spatial attention, multi-scale depthwise blocks and a dual-pooling classification head within 290,572 trainable parameters. The individual components are drawn from established literature; the contribution lies in the composition and in the parameter budget it achieves.
\item We benchmark \nameModel{} against six ImageNet-pretrained lightweight backbones under an identical data pipeline, split and evaluation protocol, showing that a model trained from scratch at 0.29M parameters exceeds every baseline while using 8.7 to 16.8 times fewer parameters and 1.3 to 8.5 times fewer multiply--accumulate operations.
\item We verify that the architecture transfers without redesign to two independently collected Bangladeshi datasets, reaching 98.71\% accuracy over 21 vegetable classes and 99.05\% accuracy over 5 radish classes.
\item We report a five-seed reproducibility study, an eleven-variant component ablation and Grad-CAM interpretability evidence, establishing that the reported accuracy is stable across initialization and that predictions rest on lesion-bearing regions.
\end{itemize}
The study is organized around five research questions:
\begin{itemize}
\item \textbf{RQ1:} Can a convolutional network trained from scratch with fewer than 0.3M parameters match ImageNet-pretrained lightweight backbones on a region-specific, multi-crop disease benchmark?
\item \textbf{RQ2:} Which architectural components of \nameModel{} contribute to its accuracy, and what parameter cost does each one carry?
\item \textbf{RQ3:} Does the architecture generalize to independently collected plant disease datasets without any change to its structure or its training recipe?
\item \textbf{RQ4:} How stable is the reported performance across random initialization and stochastic optimization?
\item \textbf{RQ5:} Do the learned representations attend to disease-bearing leaf regions rather than to background or acquisition artifacts?
\end{itemize}
The remainder of this paper is organized as follows. Section~\ref{sec:background} introduces the necessary background, Section~\ref{sec:literature} reviews related work, Section~\ref{sec:dataset} describes the construction and composition of \nameDataset, Section~\ref{sec:methodology} presents \nameModel{} and the experimental protocol, Section~\ref{sec:results} reports the results, and Section~\ref{sec:conclusion} concludes.

\section{Background}
\label{sec:background}
\begin{figure}[t]
\centering
\includegraphics[width=0.6\linewidth]{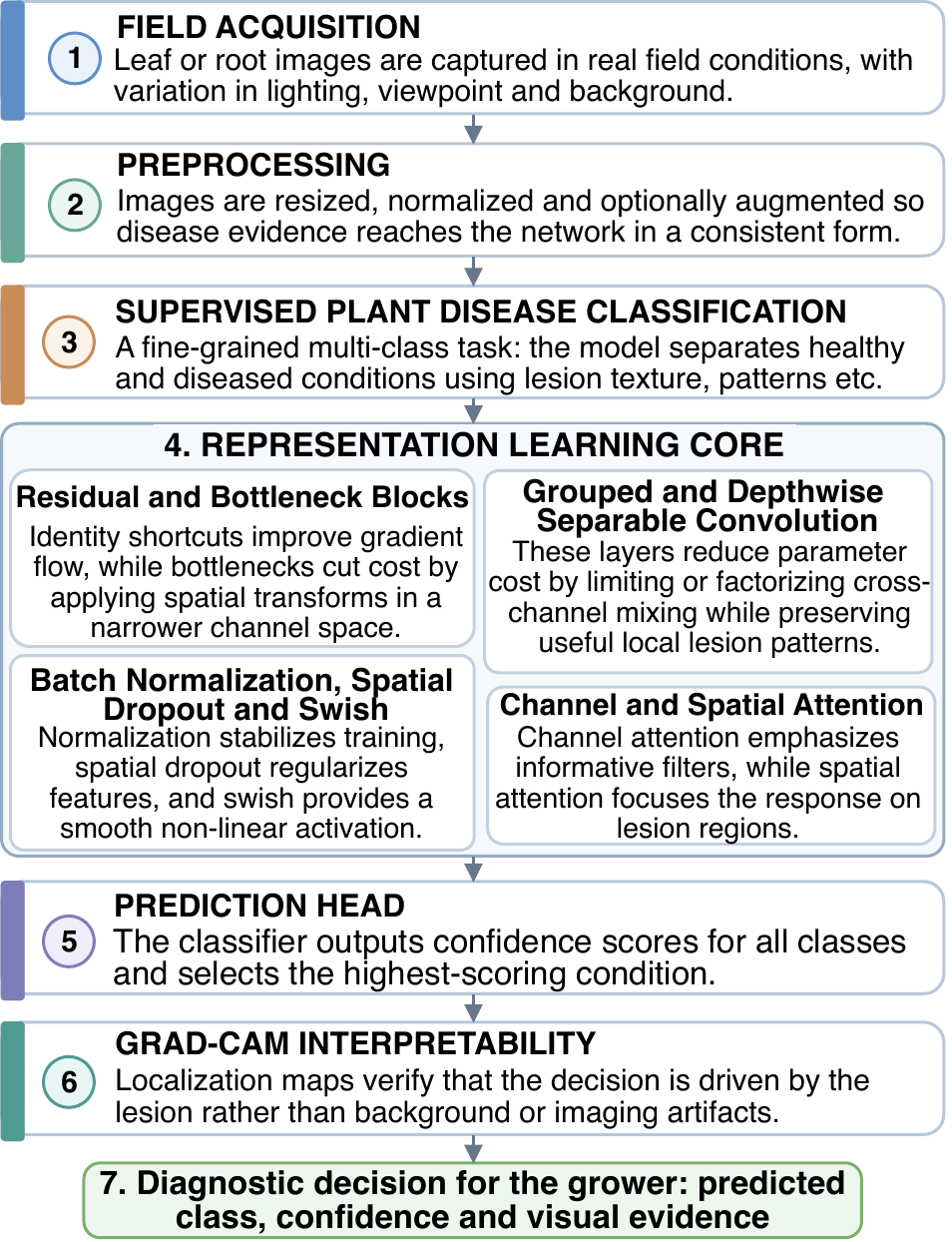}
\caption{Conceptual pipeline of an image-based plant disease diagnosis system, from field acquisition through preprocessing, representation learning and interpretability to the diagnostic decision returned to the grower.}
\label{fig:background_overview}
\end{figure}

This section introduces the concepts on which the remainder of the paper depends, at a descriptive level; the formal treatment of each mechanism, together with the notation used throughout, is deferred to Section~\ref{sec:methodology}. Figure~\ref{fig:background_overview} places the concepts within the end-to-end pipeline of an image-based plant disease diagnosis system.
\subsection{Plant Disease Classification as a Supervised Vision Task}
Plant disease classification is posed as multi-class image classification. A color photograph of a leaf or a root is presented to a trained network, which returns a confidence score for every healthy and diseased condition in the label set and assigns the image to the condition receiving the highest score. The task is fine grained rather than merely multi-class: several conditions differ only in lesion texture, in the morphology of the lesion margin or in the pattern of yellowing, while sharing leaf shape, venation and overall color envelope. This visual proximity between classes, rather than the raw number of classes, determines the representational demand placed on the network and motivates every design decision described in Section~\ref{sec:methodology}.
\subsection{Residual and Bottleneck Convolutional Blocks}
Residual learning adds a direct path from the input of a group of convolutional layers to its output, allowing the group to learn a correction to its input rather than a complete transformation of it. The direct path carries gradients back to early layers without attenuation, which permits substantial depth without the accuracy degradation that plagued earlier deep stacks~\citep{he2016deep}. The bottleneck variant narrows the representation before the spatial convolution and widens it again afterward, confining the expensive spatial operation to a reduced channel space and lowering its cost sharply. \nameModel{} adopts the bottleneck form throughout.
\subsection{Grouped and Depthwise Separable Convolution}
The weight count of a conventional convolution grows with the product of its input and output channel counts, which makes it the dominant cost in a deep network. Grouped convolution partitions the channels into disjoint groups and convolves within each group independently, dividing that cost by the number of groups~\citep{krizhevsky2012imagenet}. Depthwise separable convolution takes this partitioning to its limit, filtering each channel in isolation and then recombining the channels with a pointwise projection, the construction underlying MobileNet~\citep{howard2017mobilenets}. Both mechanisms trade a degree of cross-channel mixing for parameter economy, and \nameModel{} applies both at different depths of the network.
\subsection{Channel and Spatial Attention}
Attention modules rescale intermediate feature maps by a gate that the network learns alongside its filters. Channel attention summarizes each feature map into a single descriptor, passes the descriptors through a small perceptron and emits one multiplier per channel, which amplifies informative filters and suppresses the rest, the mechanism introduced by squeeze-and-excitation networks~\citep{hu2018squeeze}. Spatial attention instead summarizes across channels and emits one multiplier per spatial location, which concentrates response on the region carrying the evidence. The convolutional block attention module composes the two in sequence and reports consistent gains at negligible parameter cost~\citep{woo2018cbam}. Lesion classification benefits from both forms: the channel gate emphasizes texture-selective filters, and the spatial gate directs the response toward the lesion rather than the surrounding healthy tissue.
\subsection{Regularization and Activation}
Batch normalization standardizes the intermediate activations within each mini-batch, which stabilizes optimization and permits higher learning rates~\citep{ioffe2015batch}. Dropout counters co-adaptation between units by removing a random subset of them at each training step, and its structured variant spatial dropout removes entire feature maps instead, which is the appropriate form for convolutional tensors whose neighboring activations are strongly correlated~\citep{srivastava2014dropout, tompson2015efficient}. The swish activation is smooth and non-monotonic, and it is reported to outperform the rectified linear unit on deeper convolutional stacks~\citep{ramachandran2017searching}. \nameModel{} uses all four.
\subsection{Gradient-Weighted Class Activation Mapping}
Grad-CAM produces a class-discriminative localization map by weighting the activations of a late convolutional layer according to how strongly each one influences the score of the predicted class, then overlaying the result on the input image~\citep{selvaraju2017grad}. In agricultural diagnosis its role is diagnostic rather than decorative. A model that separates classes on background soil, on illumination gradients or on artifacts of the acquisition setup will produce maps that fall away from the lesion, and this failure mode remains invisible in aggregate accuracy, surfacing only once the model meets imagery from a different collection.
\section{Literature Review}
\label{sec:literature}
We organize prior work along five research lines: transfer learning on public benchmarks, purpose-built architectures for radish, potato and pointed gourd, efficient architectures for on-device inference, region-specific data collection, and interpretability. Table~\ref{tab:literature} summarizes the studies most directly comparable to ours.
\subsection{Transfer Learning on Public Benchmarks}
The dominant methodology fine-tunes an ImageNet-pretrained backbone on PlantVillage. \citet{hassan2021identification} evaluated InceptionV3, InceptionResNetV2, MobileNetV2 and EfficientNetB0 across the 38 classes of the benchmark, with EfficientNetB0 reaching 99.56\%. \citet{pandian2022plant} scaled the setting to 147,500 images over 59 classes with a 14-layer network and three augmentation strategies, reporting 99.97\%. \citet{harakannanavar2022plant} contrasted classical pipelines built on discrete wavelet transform, principal component analysis and gray-level co-occurrence matrix features against convolutional models on tomato imagery, with the convolutional model reaching 99.6\% against 88\% for a support vector machine. \citet{ali2024ensemble} pushed an ensemble of deep architectures to 99.89\% on the same benchmark. These studies establish that the PlantVillage distribution is close to saturated, yet accuracy on it carries limited information about behavior on imagery acquired under a different pathogen profile and a different acquisition protocol.
\subsection{Purpose-Built Architectures for radish and potato}
A second line designs architectures for a single crop. \citet{rozaqi2020identification}, \citet{sanjeev2021early} and \citet{barman2020comparative} each proposed a network for the three-class potato problem of early blight, late blight and healthy leaf, reporting 92\%, 96.5\% and 96.75\% respectively, all trained on PlantVillage. \citet{rashid2021multi} raised this to 99.75\% with a multi-level formulation. \citet{reis2024potato} introduced MDSCIRNet, which composes depthwise separable convolution with multi-head attention and reaches 99.24\% alone and 99.33\% when its features are passed to a support vector machine. \citet{dame2025deep} classified 4,200 smartphone potato images into three conditions at 99\% accuracy and additionally graded severity into six levels at 96\%. Radish has received markedly less attention: \citet{banerjee2023integrated} reported 92\% over five radish diseases with a convolutional and support-vector hybrid, and \citet{alam2025deep} applied a modified Mask R-CNN to radish organ segmentation rather than to disease. Pointed gourd, to the best of our knowledge, has not been treated in this literature at all.
\subsection{Efficient Architectures for On-Device Inference}
Deployment on grower-held hardware has motivated a family of compact backbones that falls into three design lineages. The purely convolutional line comprises GhostNetV2, which generates part of its feature maps through cheap linear operations~\citep{tang2022ghostnetv2}, RepGhostNet, which reparameterizes the same idea for hardware efficiency~\citep{chen2022repghost}, ConvNeXt, which modernizes the plain convolutional stack with large kernels and inverted bottlenecks~\citep{liu2022convnet}, and MobileNetV4, whose universal inverted bottleneck is searched for Pareto-optimal latency across mobile processors and accelerators~\citep{qin2024mobilenetv4}. A second line applies structural reparameterization to hybrid designs, with FastViT removing skip connections from its token mixer to lower memory access cost~\citep{vasu2023fastvit} and RepViT transferring the block-level, macro and micro design choices of lightweight transformers back onto a pure convolutional network~\citep{wang2024repvit}. A third line reduces the cost of attention itself: EdgeNeXt and MobileViTv2 interleave convolution with separable self-attention~\citep{maaz2022edgenext, mehta2022separable}, SwiftFormer replaces quadratic attention with an additive formulation~\citep{shaker2023swiftformer}, and EfficientFormerV2 searches latency and parameter count jointly to reach MobileNet-scale budgets~\citep{li2023efficientformerv2}. Within agriculture, \citet{ahmed2026large} tried to cover deployable models of lightweight architecture in leaf disease domain and \citet{vishnoi2022detection} reduced layer count to classify apple disease at 98\% with lower storage and execution cost, and \citet{rakesh2025implementation} deployed ResNet50 and DenseNet121 on a Raspberry Pi 4B for root crop classification at 99.60\% and 97.60\%. These architectures establish that competitive accuracy is attainable in the one to five million parameter range, however all of them are pretrained on ImageNet and none is designed against the texture-driven discrimination that lesion classification demands.
\subsection{Region-Specific Data Collection}
A smaller line of work addresses the data gap directly. \citet{waldchen2018automated} argued that automated species and disease identification requires collections representative of the target climatic region. \citet{hasan2024comprehensive} released 25 categories of Bangladeshi vegetable leaf imagery and \citet{hasan2025smartphone} followed with 2,801 radish leaf images across one healthy and four diseased classes. \citet{qin2021attention} evaluated ResNet variants on 15,207 field images covering 201 species, reporting 91.83\% to 92.95\%. These releases confirm the value of native imagery, yet each covers a single crop family, none reports formal agronomist validation of the class assignments, and none is paired with an architecture designed for the resulting distribution.
\subsection{Interpretability in Agricultural Diagnosis}
Interpretability has entered the field as a trust requirement. \citet{alhammad2025deep} combined transfer learning with Grad-CAM on 2,152 potato images, reaching 98\% test accuracy with visual explanations of each decision. \citet{paul2024study} applied LIME and SHAP to a deep ensemble over four potato diseases collected in West Bengal. These studies treat interpretability qualitatively, presenting selected heatmaps without a quantitative agreement measure against expert annotation.
\subsection{Research Gap}
Across these five lines, existing methods satisfy at most two of the three requirements of our setting: an expert-validated dataset representative of the local cropping system, a parameter budget compatible with low-cost hardware, and evidence that accuracy survives a change of dataset. Transfer learning studies satisfy none of the three, purpose-built potato architectures satisfy the second alone, efficient backbones satisfy the second while relying on ImageNet pretraining, and region-specific data releases satisfy the first while offering no matching model. \nameDataset{} and \nameModel{} are constructed to satisfy all three jointly.
\begin{table*}[t]
\centering
\caption{Overview of deep learning approaches to radish and potato disease classification. Accuracies are quoted from the original publications and were obtained under differing splits and protocols.}
\label{tab:literature}
\small
\begin{tabular}{lp{.5cm}rllr}
\toprule
Study & Crop & Classes & Dataset & Method & Accuracy (\%) \\
\midrule
\citet{rozaqi2020identification} & Potato & 3 & PlantVillage & CNN & 92.00 \\
\citet{sanjeev2021early} & Potato & 3 & PlantVillage & Feed-forward network & 96.50 \\
\citet{barman2020comparative} & Potato & 3 & PlantVillage & Self-built CNN & 96.75 \\
\citet{rashid2021multi} & Potato & 3 & PlantVillage & Multi-level CNN & 99.75 \\
\citet{lee2020health} & Potato & 3 & PlantVillage & CNN & 99.00 \\
\citet{jha2024deep} & Potato & 3 & Field-collected & Stacking ensemble & 98.86 \\
\citet{reis2024potato} & Potato & 4 & Public composite & MDSCIRNet with SVM & 99.33 \\
\citet{alam2025deep} & Radish & 2 & Field-collected & Mask R-CNN & 96.30 \\
\citet{banerjee2023integrated} & Radish & 5 & Field-collected & CNN with SVM & 92.00 \\
\citet{ullah2023effective} & Multiple & 8 & PlantVillage & DeepPlantNet & 98.49 \\
\citet{geetharamani2019identification} & Multiple & 39 & PlantVillage & Nine-layer CNN & 96.46 \\
\citet{zhang2020deep} & Tomato & 4 & AIChallenger & Improved Faster R-CNN & 97.10 \\
\citet{qin2021attention} & Multiple & 201 & Field-collected & Res2Net-101 & 92.95 \\
\citet{shafik2025deep} & Multiple & 15 & TPPD  & ResNet-9 & 97.40 \\
\citet{nawaz2026potatoguardnet} & Potato & 3 & PlantVillage  & PotatoGuardNet & 99.41 \\
\citet{sanchez2026development} & Potato & 2 & Field-collected   & ResNet50 & 93.00 \\
\citet{dame2025deep} & Potato & 3 & Field-collected   &  CNN & 99.00 \\
\citet{sinamenye2025potato} & Potato & 7 & Field-collected   &  Hybrid deep learning & 85.06 \\
\citet{quoc2025vision} & Radish & 5 & Bangladesh   &  SCOLD & 94.37 \\
\citet{ji2024implementing} & Radish & 6 & Field-collected   &  Hybrid CNN–Transformer & 91.00 \\

\midrule
\textbf{This work} & \textbf{Radish, pointed gourd, potato} & \textbf{12} & \textbf{\nameDataset} & \textbf{\nameModel{} (0.29M params)} & \textbf{99.52} \\
\bottomrule
\end{tabular}%

\end{table*}
\section{The \nameDataset{} Dataset}
\label{sec:dataset}
\begin{figure}[t]
\centering
\includegraphics[width=0.6\linewidth]{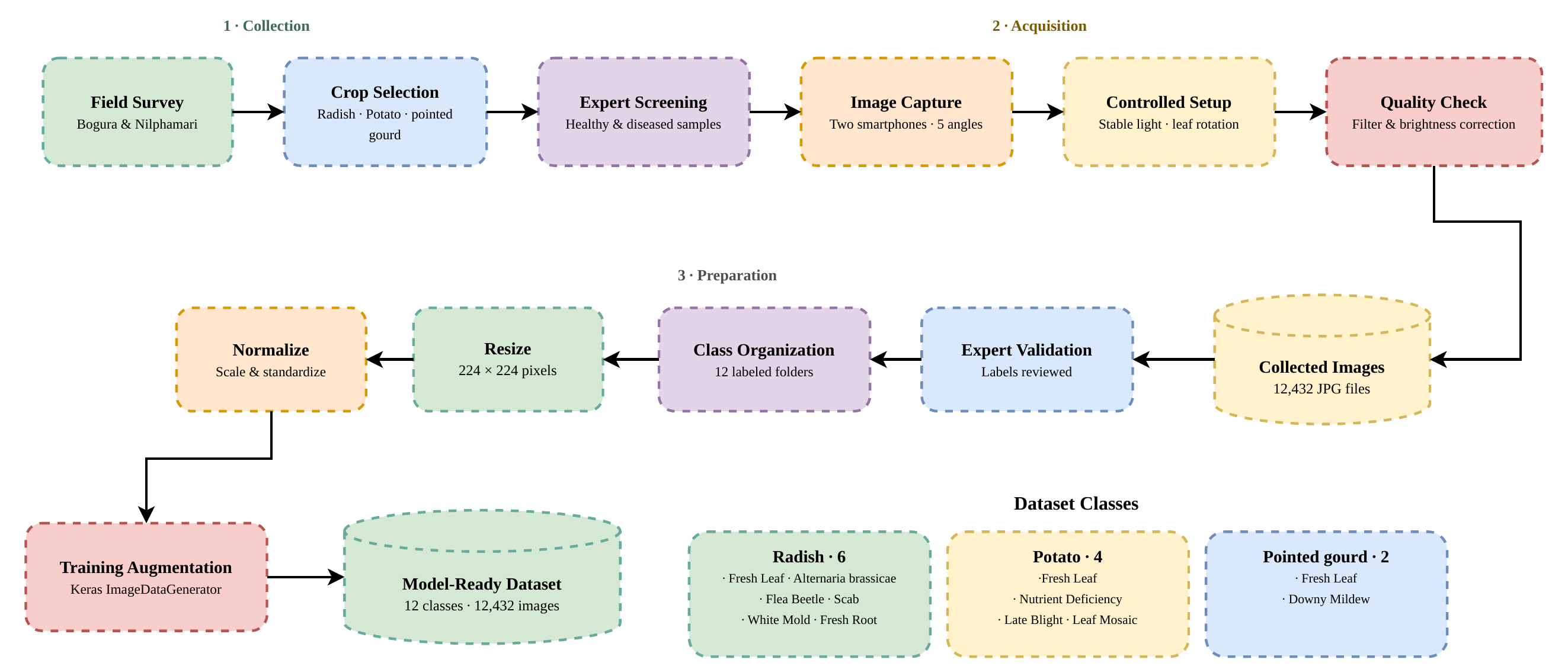}
\caption{Construction workflow of \nameDataset, from agronomist-guided field survey and multi-angle acquisition through curation, expert validation of the class assignment and stratified partition into training, validation and test splits.}
\label{fig:dataset_pipeline}
\end{figure}

\nameDataset{} is an expert-validated collection of 12,432 field images covering 12 classes drawn from three vegetable crops of northern Bangladesh, each represented in healthy and diseased condition. This section reports the agronomic rationale for the crop selection, the acquisition and curation protocol, the expert validation procedure, the composition of the release and the properties that distinguish it from existing collections. Figure~\ref{fig:dataset_pipeline} summarizes the workflow from field survey to released split.
\subsection{Crop Selection and Survey Regions}
Vegetable production occupies 2.57\% of the arable land of Bangladesh and yields 3.73 million tons annually, a return per hectare that exceeds that of the staple cereals~\citep{haque2021vegetable}. Within this sector the Bogura and Nilphamari districts are among the most intensive winter vegetable regions of the country~\citep{fe2016bogra}. Consultation with local agronomists identified radish (\textit{Raphanus sativus}), potato (\textit{Solanum tuberosum}) and pointed gourd (\textit{Trichosanthes dioica}, known locally as potol) as the three crops that combine the largest cultivated area with the highest reported disease-driven loss in these districts. Radish and potato are represented in the existing literature, however almost always through a single foliar presentation and on imagery acquired outside South Asia; pointed gourd, to the best of our knowledge, is absent from the published disease classification literature entirely.
\subsection{Acquisition Protocol}
Field surveys were conducted across several agricultural sites in the two districts over the winter cropping season. Local agronomists accompanied each survey and identified the diseased plots in advance, which ensured that the sampled plants carried the pathology of interest rather than an incidental stress symptom. Specimens were harvested from the identified plots and cataloged against the plot record before imaging. Two consumer smartphones, a POCO M3 and an OPPO F16, served as the acquisition devices, which reflects the sensor class a grower would realistically use at inference time. Each specimen was photographed from five angles under a controlled illumination arrangement that suppresses direct sunlight, preserves color fidelity and avoids the saturation that removes lesion texture. Specimen position was rotated manually between exposures to capture the adaxial surface, the abaxial surface and the lesion margin. Images were stored as JPEG at the native resolution of the capture device, a format chosen for its storage economy and its universal support in mobile acquisition pipelines.
\subsection{Curation and Expert Validation}
Raw captures were screened for motion blur, severe defocus and duplicate framing, and the surviving images were assigned to one of the 12 classes. Each class assignment was then reviewed by practicing agronomists, who confirmed the pathology, the affected organ and the severity band on a per-image basis; images on which the reviewers disagreed were removed rather than resolved by majority. The signed validation record accompanies the release. This step distinguishes \nameDataset{} from the region-specific collections currently available, which report the label taxonomy without documenting an independent verification of the assignments~\citep{hasan2024comprehensive, hasan2025smartphone}. Filtering was additionally applied to normalize brightness across acquisition sessions, and no synthetic or generative augmentation entered the released images; all augmentation reported in Section~\ref{sec:methodology} is applied at training time only.
\subsection{Composition}
\begin{figure}[t]
\centering
\includegraphics[width=0.5\linewidth]{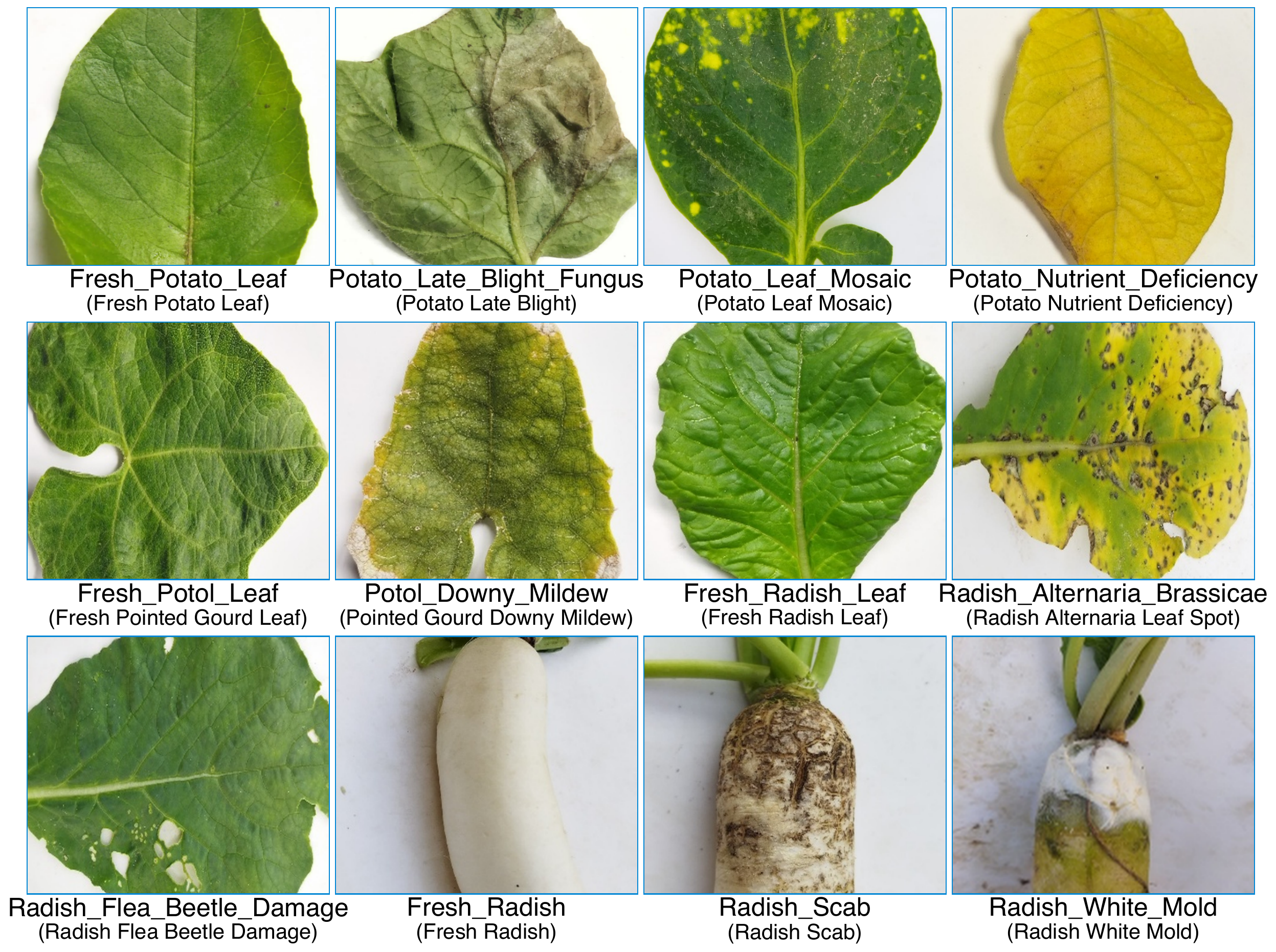}
\caption{Representative images from each of the 12 \nameDataset{} classes, arranged by crop, with healthy presentations in the first column of each crop group.}
\label{fig:dataset_overview}
\end{figure}

\begin{figure}[t]
\centering
\includegraphics[width=0.6\linewidth]{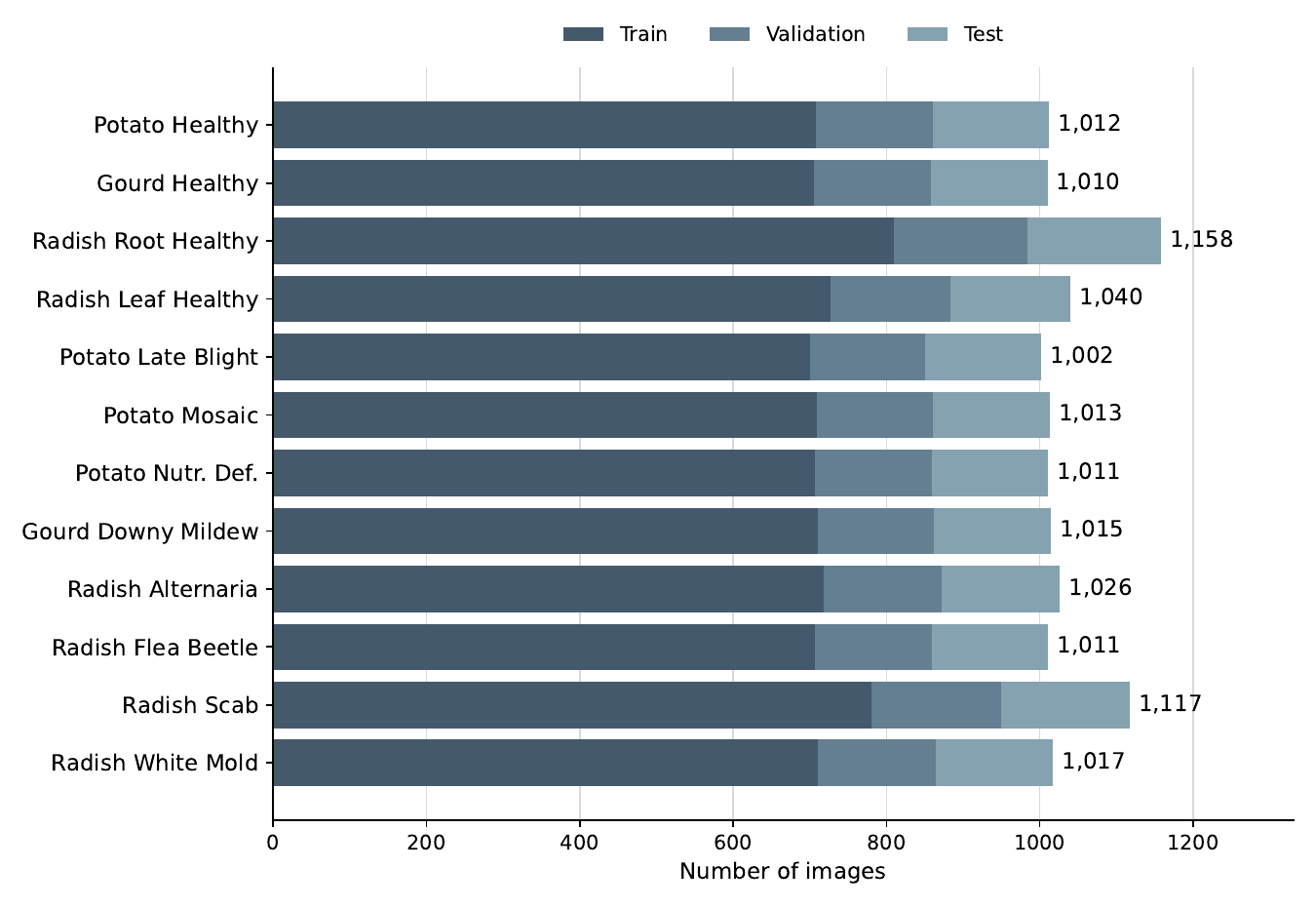}
\caption{Per-class image counts in \nameDataset{} across the training, validation and test partitions. The maximum imbalance ratio between the largest and the smallest class is 1.16.}
\label{fig:dataset_distribution}
\end{figure}

The release contains 12,432 images across 12 classes. Radish contributes six classes: two healthy presentations covering the root and the leaf, together with \textit{Alternaria brassicae} blight, flea beetle damage, scab and white mold. Potato contributes four classes: a healthy leaf, late blight fungus, leaf mosaic and nutrient deficiency. Pointed gourd contributes two classes: a healthy leaf and downy mildew. The taxonomy therefore spans four distinct etiologies, namely fungal infection, viral infection, insect damage and abiotic nutrient disorder, which is a wider causal range than single-pathogen benchmarks provide and which forces the model to separate visually similar chlorotic patterns of different origin. Inclusion of the radish root alongside the radish leaf further requires the model to operate across two organ morphologies within one label space. Table~\ref{tab:dataset_composition} lists the per-class composition and Figure~\ref{fig:dataset_distribution} shows the distribution over the three splits. Class counts range from 1,002 to 1,158 images, giving a maximum imbalance ratio of 1.16 and removing the need for resampling or class-weighted loss. Figure~\ref{fig:dataset_overview} presents representative images for each class.
\begin{table*}[t]
\centering
\caption{Composition of \nameDataset. Counts follow a stratified 70/15/15 partition applied within each class.}
\label{tab:dataset_composition}
\small
\begin{tabular}{llrrrrr}
\toprule
Crop & Class & Condition & Train & Val. & Test & Total \\
\midrule
Potato & Potato Healthy & Healthy & 708 & 152 & 152 & 1,012 \\
Pointed gourd & Gourd Healthy & Healthy & 706 & 152 & 152 & 1,010 \\
Radish & Radish Root Healthy & Healthy & 810 & 174 & 174 & 1,158 \\
Radish & Radish Leaf Healthy & Healthy & 727 & 156 & 157 & 1,040 \\
Potato & Potato Late Blight & Diseased & 700 & 151 & 151 & 1,002 \\
Potato & Potato Mosaic & Diseased & 709 & 152 & 152 & 1,013 \\
Potato & Potato Nutrient Deficiency & Diseased & 707 & 152 & 152 & 1,011 \\
Pointed gourd & Gourd Downy Mildew & Diseased & 710 & 152 & 153 & 1,015 \\
Radish & Radish \textit{Alternaria brassicae} & Diseased & 718 & 154 & 154 & 1,026 \\
Radish & Radish Flea Beetle Damage & Diseased & 707 & 152 & 152 & 1,011 \\
Radish & Radish Scab & Diseased & 781 & 168 & 168 & 1,117 \\
Radish & Radish White Mold & Diseased & 711 & 153 & 153 & 1,017 \\
\midrule
\multicolumn{3}{l}{Total} & 8,694 & 1,868 & 1,870 & 12,432 \\
\bottomrule
\end{tabular}
\end{table*}
\subsection{Partitioning}
The collection is partitioned once into 8,694 training, 1,868 validation and 1,870 test images under a stratified 70/15/15 scheme, and this fixed partition is reused without modification by every experiment in Section~\ref{sec:results}. Partitioning is applied at the specimen level rather than at the image level, which prevents the five exposures of one physical leaf from being distributed across splits and removes the optimistic bias that image-level partitioning introduces into multi-view collections. The test split is touched once per configuration, after model selection has been completed on the validation split.
\subsection{Comparison with Existing Collections}
Table~\ref{tab:dataset_comparison} positions \nameDataset{} against the public collections most closely related to it. PlantVillage remains the largest, however its imagery is laboratory acquired and covers no crop native to the target cropping system. The two Bangladeshi releases of \citet{hasan2024comprehensive} and \citet{hasan2025smartphone} are region appropriate, yet the first covers foliar presentations only and the second is restricted to radish. \nameDataset{} is, to the best of our knowledge, the first collection to combine multi-crop coverage of this cropping system, both foliar and root presentations, four distinct disease etiologies and documented per-image expert validation.
\begin{table*}[t]
\centering
\caption{\nameDataset{} against related public collections.}
\label{tab:dataset_comparison}
\small
\begin{tabular}{llrrlll}
\toprule
Dataset & Region & Images & Classes & Crops & Organs & Expert validated \\
\midrule
PlantVillage~\citep{hassan2021identification} & Not region specific & 54,305 & 38 & 14 & Leaf & Not reported \\
VegNet-BD~\citep{hasan2024comprehensive} & Bangladesh & 12,786 & 21 & 6 & Leaf & Not reported \\
RadishLeaf-BD~\citep{hasan2025smartphone} & Bangladesh & 2,801 & 5 & 1 & Leaf & Not reported \\
\citet{qin2021attention} & China & 15,207 & 201 & 201 & Leaf & Not reported \\
\midrule
\textbf{\nameDataset{} (ours)} & \textbf{Bangladesh} & \textbf{12,432} & \textbf{12} & \textbf{3} & \textbf{Leaf and root} & \textbf{Yes, per image} \\
\bottomrule
\end{tabular}
\end{table*}

\section{Methodology}
\label{sec:methodology}
Section~\ref{sec:background} described the mechanisms on which \nameModel{} rests in words; this section states them formally, derives the network block by block against the architecture of Figure~\ref{fig:architecture}, and fixes the training and evaluation protocol. Every symbol introduced here is collected in Table~\ref{tab:notation}, which is self-contained and carries the full notation of the paper.
\begin{table*}[t]
\centering
\caption{List of Notations.}
\label{tab:notation}
\scriptsize
\setlength{\tabcolsep}{7pt}
\setlength{\extrarowheight}{0pt}
\renewcommand{\arraystretch}{1.25}
\begin{tabular}{|c|p{0.3\textwidth}|c|p{0.3\textwidth}|}
\hline
\textbf{Symbol} & \textbf{Description} & \textbf{Symbol} & \textbf{Description} \\
\hline
\hline
$\mathbf{X}$ & Raw RGB input image, $\mathbf{X} \in \mathbb{R}^{H \times W \times 3}$ & $\sigma(\cdot), \rho(\cdot)$ & Sigmoid and rectified linear activation \\
\hline
$H, W$ & Input height and width, both set to 224 & $\mathrm{BN}(\cdot)$ & Batch normalization \\
\hline
$H', W'$ & Height and width of an intermediate feature map & $\mathrm{GAP}, \mathrm{GMP}$ & Global average and global maximum pooling \\
\hline
$C$ & Channel count at the input of a block & $\mathbf{W}_{\text{stem}}$ & Kernel of the strided stem convolution \\
\hline
$\widetilde{\mathbf{X}}$ & Rescaled image with entries in $[0,1]$ & $\mathbf{W}_{r1}, \mathbf{W}_{r2}, \mathbf{W}_{r3}$ & Kernels of the grouped residual bottleneck \\
\hline
$\widehat{\mathbf{X}}$ & Augmented network input & $\mathbf{W}_{0}, \mathbf{W}_{1}$ & Weights of the shared attention perceptron \\
\hline
$K$ & Number of classes, $K = 12$ on \nameDataset & $\mathbf{W}_{s}$ & Kernel of the spatial attention gate \\
\hline
$N$ & Number of images in a split & $\mathbf{W}_{p}, \mathbf{W}_{q}$ & Pointwise kernels of the depthwise block \\
\hline
$\mathcal{D}$ & Labeled collection $\{(\mathbf{X}_{i}, \mathbf{y}_{i})\}_{i=1}^{N}$ & $\mathbf{W}_{d3}, \mathbf{W}_{d5}$ & Depthwise kernels at size 3 and size 5 \\
\hline
$\mathbf{y}, \widehat{\mathbf{y}}$ & One-hot label and predicted class distribution & $\mathbf{W}_{f1}, \mathbf{W}_{f2}, \mathbf{W}_{o}$ & Kernels of the classification head \\
\hline
$\Delta^{K-1}$ & Probability simplex over the $K$ classes & $\mathbf{M}_{c}$ & Channel attention vector, $\mathbf{M}_{c} \in \mathbb{R}^{1 \times 1 \times F}$ \\
\hline
$\boldsymbol{\theta}$ & Trainable parameters of the network & $\mathbf{M}_{s}$ & Spatial attention map, $\mathbf{M}_{s} \in \mathbb{R}^{H' \times W' \times 1}$ \\
\hline
$f_{\boldsymbol{\theta}}$ & Network mapping an image to $\Delta^{K-1}$ & $\mathbf{z}_{\text{avg}}, \mathbf{z}_{\text{max}}$ & Globally averaged and maximized channel descriptors \\
\hline
$\mathcal{A}_{\tau}$ & Stochastic training-time augmentation operator & $r$ & Channel attention reduction ratio, $r = 16$ \\
\hline
$\mathbf{U}^{(l)}$ & Input tensor of stage $l$ & $\mathcal{S}(\cdot)$ & Shortcut mapping, identity or $1 \times 1$ projection \\
\hline
$\mathbf{Z}^{(i)}$ & $i$-th intermediate tensor inside a residual block & $\mathbf{P}$ & Pointwise-expanded tensor inside a depthwise block \\
\hline
$\mathbf{Y}$ & Output tensor of an Enhanced Residual Block & $\mathbf{D}_{3}, \mathbf{D}_{5}$ & Depthwise responses at kernel size 3 and 5 \\
\hline
$F_{l}$ & Output channel width of stage $l$ & $\mathbf{Q}$ & Fused output of a Multi-Scale Depthwise Block \\
\hline
$g_{l}$ & Number of convolution groups in stage $l$ & $p_{s}$ & Spatial dropout rate of a stage \\
\hline
$k, s$ & Convolution kernel size and stride & $p_{d}$ & Dense dropout rate, $p_{d} = 0.3$ \\
\hline
$\circledast_{k,s}$ & Convolution with kernel size $k$ and stride $s$ & $\mathbf{v}$ & Dual-pooled feature vector, $\mathbf{v} \in \mathbb{R}^{2F_{4}}$ \\
\hline
$\circledast_{k,g}$ & Grouped convolution over $g$ groups & $\mathbf{h}_{1}, \mathbf{h}_{2}$ & Hidden activations of the classification head \\
\hline
$\circledast_{k}^{\text{dw}}$ & Depthwise convolution with kernel size $k$ & $\mathcal{L}$ & Total training objective \\
\hline
$\odot$ & Broadcast Hadamard product & $\lambda$ & Weight decay coefficient, $\lambda = 10^{-5}$ \\
\hline
$[\,\cdot\,;\,\cdot\,]$ & Concatenation along the channel axis & $\eta_{t}, \mathbf{m}_{t}$ & Learning rate and momentum buffer at step $t$ \\
\hline
$\lVert \cdot \rVert_{F}$ & Frobenius norm of a kernel & $\mu, B, E$ & Momentum 0.9, batch size 32, epoch cap 300 \\
\hline
$\delta(\cdot)$ & Swish activation, $\delta(x) = x\,\sigma(x)$ & $\Omega(\cdot)$ & Parameter count of a layer or a block \\
\hline
$\mathbf{A}_{k}$ & $k$-th activation map of the last convolutional tensor & $\alpha_{k}^{c}, \mathbf{L}^{c}$ & Grad-CAM channel weight and localization map for class $c$ \\
\hline
\end{tabular}
\end{table*}
\subsection{Problem Formulation}
Let $\mathcal{D} = \{(\mathbf{X}_{i}, \mathbf{y}_{i})\}_{i=1}^{N}$ denote a labeled collection in which $\mathbf{X}_{i} \in \mathbb{R}^{H \times W \times 3}$ is an RGB image of a leaf or a root and $\mathbf{y}_{i} \in \{0,1\}^{K}$ is a one-hot encoding over the $K = 12$ classes of \nameDataset. We seek parameters $\boldsymbol{\theta}$ of a network $f_{\boldsymbol{\theta}}$ that maps an image onto the probability simplex $\Delta^{K-1} = \{\mathbf{p} \in \mathbb{R}^{K} : p_{c} \geq 0, \sum_{c} p_{c} = 1\}$ and minimizes the expected categorical cross-entropy over the data distribution. The design constraint distinguishing this work from the standard formulation is a budget on $|\boldsymbol{\theta}|$: the model must remain below 0.3M trainable parameters while matching backbones one to two orders of magnitude larger.
\subsection{Preprocessing and Augmentation}
Images are decoded, resized to $224 \times 224$ by bilinear interpolation and rescaled to unit range,
\begin{equation}
\widetilde{\mathbf{X}} = \frac{1}{255}\,\mathbf{X}.
\label{eq:rescale}
\end{equation}
Augmentation is embedded inside the network graph as a stochastic operator $\mathcal{A}_{\tau}$ that is active during training and reduces to the identity at inference,
\begin{equation}
\widehat{\mathbf{X}} =
\begin{cases}
\mathcal{A}_{\tau}(\widetilde{\mathbf{X}}), & \text{training},\\[2pt]
\widetilde{\mathbf{X}}, & \text{inference}.
\end{cases}
\label{eq:augment}
\end{equation}
The operator composes random rotation within $\pm 15^{\circ}$, random zoom within $\pm 10\%$, random translation within $\pm 10\%$ of each spatial extent and random contrast jitter of $\pm 10\%$, each with reflective boundary fill and a fixed seed. Horizontal and vertical flips are deliberately excluded: lesion distribution on a leaf is not mirror symmetric with respect to the petiole, and flipping was found to degrade the separation of the two radish healthy presentations. Embedding $\mathcal{A}_{\tau}$ in the graph rather than in the input pipeline guarantees that the deployed model and the trained model share one preprocessing definition.
\subsection{Architecture Overview}
\begin{figure*}[t]
\centering
\includegraphics[width=\textwidth]{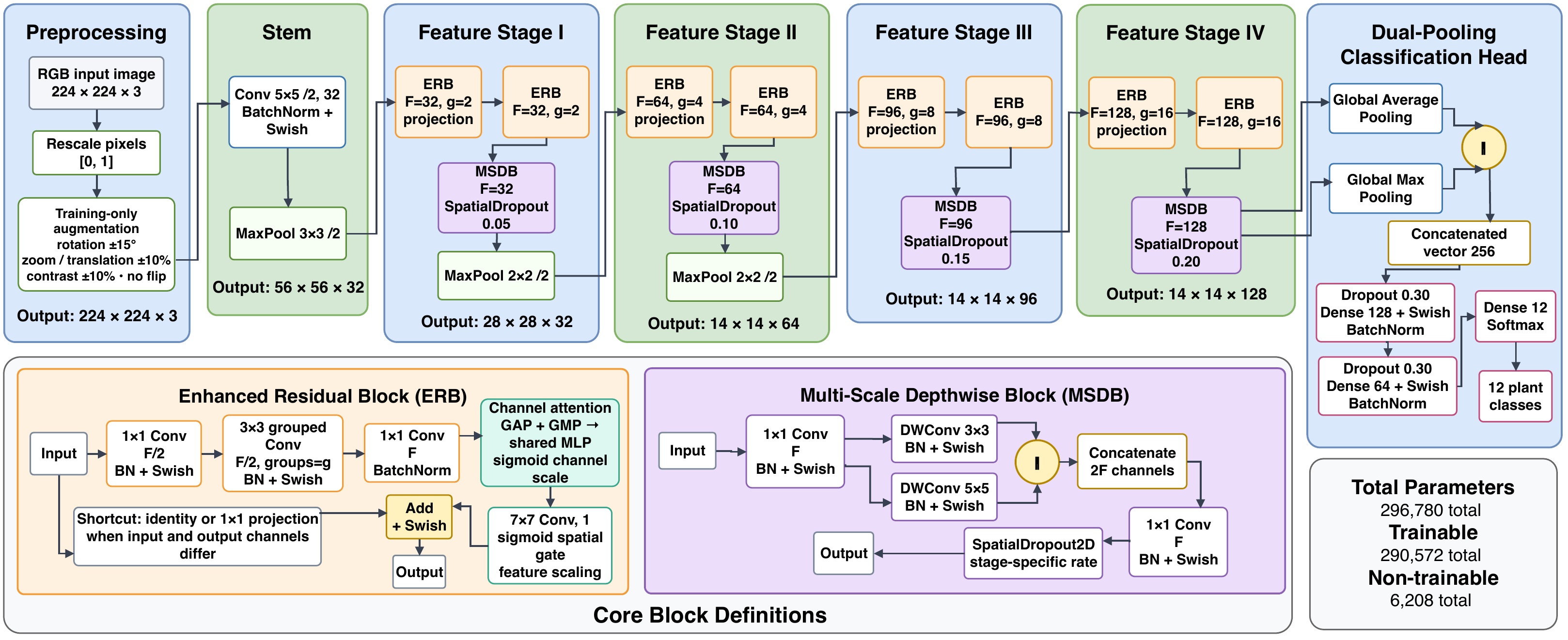}
\caption{Architecture of \nameModel. The upper row traces the forward path from preprocessing through the stem, four feature stages and the dual-pooling classification head, annotating each stage with its channel width $F_{l}$, group count $g_{l}$ and output resolution. The lower row expands the two recurring blocks: the Enhanced Residual Block, which applies a grouped bottleneck followed by sequential channel and spatial attention, and the Multi-Scale Depthwise Block, which fuses $3 \times 3$ and $5 \times 5$ depthwise responses.}
\label{fig:architecture}
\end{figure*}

Figure~\ref{fig:architecture} presents \nameModel{} in full. The network comprises a preprocessing front end, a strided stem, four feature stages and a dual-pooling classification head. Each feature stage $l \in \{1,2,3,4\}$ pairs two Enhanced Residual Blocks (ERB) with one Multi-Scale Depthwise Block (MSDB), and is characterized by a channel width $F_{l}$ and a group count $g_{l}$ with $(F_{l}, g_{l})$ taking the values $(32, 2)$, $(64, 4)$, $(96, 8)$ and $(128, 16)$. The group count is doubled together with the channel width at every stage, which holds the cost of the spatial convolution approximately constant as the representation deepens. Spatial resolution follows $224 \rightarrow 56 \rightarrow 28 \rightarrow 14 \rightarrow 14$, with downsampling performed by the stem and by the max-pooling layers that terminate stages one and two; stages three and four operate at a fixed $14 \times 14$ grid, which preserves the spatial detail on which lesion margin discrimination depends. The complete network holds 296,780 parameters, of which 290,572 are trainable and 6,208 are the non-trainable running statistics of the batch normalization layers.
\subsection{Stem}
The stem reduces the input resolution by a factor of four before any residual computation, which removes the dominant cost of full-resolution processing. It applies a strided $5 \times 5$ convolution with 32 filters, batch normalization and swish activation, followed by strided max pooling,
\begin{equation}
\mathbf{U}^{(1)} = \mathrm{MaxPool}_{3,2}\!\left( \delta\!\left( \mathrm{BN}\!\left( \mathbf{W}_{\text{stem}} \circledast_{5,2} \widehat{\mathbf{X}} \right) \right) \right),
\label{eq:stem}
\end{equation}
where $\circledast_{k,s}$ denotes convolution with kernel size $k$ and stride $s$, $\mathrm{MaxPool}_{k,s}$ denotes max pooling with window $k$ and stride $s$, and $\mathbf{U}^{(1)} \in \mathbb{R}^{56 \times 56 \times 32}$. A $5 \times 5$ kernel is preferred over the more common $3 \times 3$ at this depth since the receptive field it establishes covers a full lesion nucleus at the native scale of the imagery.
\subsection{Enhanced Residual Block}
The ERB is the primary representational unit and corresponds to the orange panel of Figure~\ref{fig:architecture}. It realizes the residual bottleneck of Section~\ref{sec:background} with grouped spatial convolution and two attention gates. Given an input $\mathbf{U}$ with $C$ channels and a target width $F$, the block first contracts to $F/2$ channels with a pointwise convolution, applies the spatial convolution in the contracted space with $g$ groups, and expands back to $F$ channels,
\begin{align}
\mathbf{Z}^{(1)} &= \delta\!\left( \mathrm{BN}\!\left( \mathbf{W}_{r1} \circledast_{1,1} \mathbf{U} \right) \right), \label{eq:erb1}\\
\mathbf{Z}^{(2)} &= \delta\!\left( \mathrm{BN}\!\left( \mathbf{W}_{r2} \circledast_{3,g} \mathbf{Z}^{(1)} \right) \right), \label{eq:erb2}\\
\mathbf{Z}^{(3)} &= \mathrm{BN}\!\left( \mathbf{W}_{r3} \circledast_{1,1} \mathbf{Z}^{(2)} \right), \label{eq:erb3}
\end{align}
with $\mathbf{W}_{r1} \in \mathbb{R}^{1 \times 1 \times C \times F/2}$, $\mathbf{W}_{r2} \in \mathbb{R}^{3 \times 3 \times (F/2g) \times F/2}$ and $\mathbf{W}_{r3} \in \mathbb{R}^{1 \times 1 \times F/2 \times F}$. The joint effect of the bottleneck and of grouping is multiplicative. A dense $3 \times 3$ convolution at width $F$ carries $9F^{2}$ weights, whereas Eq.~\eqref{eq:erb2} carries
\begin{equation}
\Omega\!\left( \mathbf{W}_{r2} \right) = \frac{9F^{2}}{4g},
\label{eq:groupcost}
\end{equation}
a reduction of $4g$, which reaches a factor of 64 in the final stage where $g = 16$. This single design choice accounts for the majority of the parameter economy of the network.
The expanded tensor is then gated along the channel axis. Two global descriptors are formed by average and maximum pooling, both are passed through a perceptron whose weights are shared across the two paths, and the summed responses are squashed to $(0,1)$,
\begin{align}
\mathbf{z}_{\text{avg}} &= \mathrm{GAP}\!\left( \mathbf{Z}^{(3)} \right), \quad \mathbf{z}_{\text{max}} = \mathrm{GMP}\!\left( \mathbf{Z}^{(3)} \right), \label{eq:desc}\\
\mathbf{M}_{c} &= \sigma\!\left( \mathbf{W}_{1}^{\top}\rho\!\left( \mathbf{W}_{0}^{\top}\mathbf{z}_{\text{avg}} \right) + \mathbf{W}_{1}^{\top}\rho\!\left( \mathbf{W}_{0}^{\top}\mathbf{z}_{\text{max}} \right) \right), \label{eq:chatt}\\
\mathbf{Z}^{(4)} &= \mathbf{M}_{c} \odot \mathbf{Z}^{(3)}, \label{eq:chapply}
\end{align}
where $\mathrm{GAP}$ and $\mathrm{GMP}$ reduce a tensor over its spatial axes, $\mathbf{W}_{0} \in \mathbb{R}^{F \times F/r}$ and $\mathbf{W}_{1} \in \mathbb{R}^{F/r \times F}$ with reduction ratio $r = 16$, both without bias, and $\odot$ broadcasts the per-channel multiplier across space. Weight sharing across the two pooling paths halves the parameter cost of the module relative to independent branches, and the two descriptors are complementary: average pooling reports the mean activation of a filter over the organ, while maximum pooling reports its peak response, which is the more informative statistic when a lesion occupies a small fraction of the frame.
A spatial gate follows. A single $7 \times 7$ convolution collapses the gated tensor to one channel and produces a per-location multiplier,
\begin{align}
\mathbf{M}_{s} &= \sigma\!\left( \mathbf{W}_{s} \circledast_{7,1} \mathbf{Z}^{(4)} \right), \label{eq:spatt}\\
\mathbf{Z}^{(5)} &= \mathbf{M}_{s} \odot \mathbf{Z}^{(4)}. \label{eq:spapply}
\end{align}
The large kernel is intentional: the gate must integrate context well beyond a lesion boundary in order to separate a genuine necrotic region from a shadow or a soil speckle of comparable local appearance. Ordering the two gates channel first and spatial second follows the convolutional block attention module~\citep{woo2018cbam}, in which this order is reported to outperform the reverse and the parallel arrangement.
The block closes with the residual addition,
\begin{equation}
\mathbf{Y} = \delta\!\left( \mathbf{Z}^{(5)} + \mathcal{S}(\mathbf{U}) \right),
\label{eq:residual}
\end{equation}
where $\mathcal{S}$ is the identity when the input and output widths agree and a batch-normalized $1 \times 1$ projection otherwise. Placing the activation after the addition rather than before it preserves an unmodified gradient path from the block output to the block input.
\subsection{Multi-Scale Depthwise Block}
The MSDB, shown in the purple panel of Figure~\ref{fig:architecture}, terminates every stage and supplies scale diversity that the fixed $3 \times 3$ kernel of the ERB cannot provide on its own. Lesion size varies substantially within a single class, from the pinhole punctures of flea beetle damage to the confluent necrotic sectors of late blight, and a single receptive field resolves only one end of that range. The block expands the input pointwise, applies two depthwise convolutions of different kernel size in parallel, concatenates their responses and projects back,
\begin{align}
\mathbf{P} &= \delta\!\left( \mathrm{BN}\!\left( \mathbf{W}_{p} \circledast_{1,1} \mathbf{Y} \right) \right), \label{eq:msdb1}\\
\mathbf{D}_{3} &= \delta\!\left( \mathrm{BN}\!\left( \mathbf{W}_{d3} \circledast_{3}^{\text{dw}} \mathbf{P} \right) \right), \label{eq:msdb2}\\
\mathbf{D}_{5} &= \delta\!\left( \mathrm{BN}\!\left( \mathbf{W}_{d5} \circledast_{5}^{\text{dw}} \mathbf{P} \right) \right), \label{eq:msdb3}\\
\mathbf{Q} &= \delta\!\left( \mathrm{BN}\!\left( \mathbf{W}_{q} \circledast_{1,1} \left[ \mathbf{D}_{3} ; \mathbf{D}_{5} \right] \right) \right), \label{eq:msdb4}\\
\mathbf{U}^{(l+1)} &= \mathrm{SpatialDropout}\!\left( \mathbf{Q}, p_{s} \right), \label{eq:msdb5}
\end{align}
where $[\,\cdot\,;\,\cdot\,]$ denotes concatenation along the channel axis and $\circledast_{k}^{\text{dw}}$ a depthwise convolution that filters each channel independently. The two depthwise branches together carry only $(9 + 25)F$ weights, which is negligible against the $F^{2}$ scaling of the pointwise projections. Spatial dropout is applied at a stage-dependent rate $p_{s} \in \{0.05, 0.10, 0.15, 0.20\}$, increasing with depth, and drops whole feature maps for the reason given in Section~\ref{sec:background}~\citep{tompson2015efficient}.
\subsection{Dual-Pooling Classification Head}
The head collapses the final $14 \times 14 \times 128$ tensor through two pooling operators in parallel and concatenates the results,
\begin{equation}
\mathbf{v} = \left[ \mathrm{GMP}\!\left( \mathbf{U}^{(5)} \right) ; \mathrm{GAP}\!\left( \mathbf{U}^{(5)} \right) \right] \in \mathbb{R}^{256}.
\label{eq:dualpool}
\end{equation}
Global average pooling summarizes the prevalence of a pattern over the whole organ, which suits diffuse conditions such as nutrient deficiency, while global maximum pooling retains the strongest local evidence, which suits focal conditions such as scab. Retaining both removes the need to select between the two failure modes. Classification proceeds through two regularized dense layers and a softmax,
\begin{align}
\mathbf{h}_{1} &= \mathrm{BN}\!\left( \delta\!\left( \mathbf{W}_{f1}^{\top}\,\mathrm{Dropout}(\mathbf{v}, p_{d}) \right) \right), \label{eq:head1}\\
\mathbf{h}_{2} &= \mathrm{BN}\!\left( \delta\!\left( \mathbf{W}_{f2}^{\top}\,\mathrm{Dropout}(\mathbf{h}_{1}, p_{d}) \right) \right), \label{eq:head2}\\
\widehat{\mathbf{y}} &= \mathrm{softmax}\!\left( \mathbf{W}_{o}^{\top}\mathbf{h}_{2} \right), \label{eq:head3}
\end{align}
with $\mathbf{W}_{f1} \in \mathbb{R}^{256 \times 128}$, $\mathbf{W}_{f2} \in \mathbb{R}^{128 \times 64}$, $\mathbf{W}_{o} \in \mathbb{R}^{64 \times K}$ and $p_{d} = 0.3$. The head accounts for 41,536 parameters, which is 14.3\% of the trainable total.
\subsection{Objective and Optimization}
Training minimizes categorical cross-entropy with an $\ell_{2}$ penalty applied to the convolutional and dense kernels,
\begin{equation}
\mathcal{L}(\boldsymbol{\theta}) = -\frac{1}{N}\sum_{i=1}^{N}\sum_{c=1}^{K} y_{i,c}\,\log \widehat{y}_{i,c} \;+\; \lambda \sum_{\mathbf{W} \in \boldsymbol{\theta}_{\text{ker}}} \lVert \mathbf{W} \rVert_{F}^{2},
\label{eq:loss}
\end{equation}
with $\lambda = 10^{-5}$ and $\boldsymbol{\theta}_{\text{ker}} \subset \boldsymbol{\theta}$ the set of convolutional and dense kernels, which excludes the batch normalization scales and offsets. Parameters are updated by stochastic gradient descent with Nesterov momentum,
\begin{align}
\mathbf{m}_{t} &= \mu\,\mathbf{m}_{t-1} + \nabla_{\boldsymbol{\theta}}\mathcal{L}\!\left( \boldsymbol{\theta}_{t-1} - \eta_{t}\mu\,\mathbf{m}_{t-1} \right), \label{eq:mom}\\
\boldsymbol{\theta}_{t} &= \boldsymbol{\theta}_{t-1} - \eta_{t}\,\mathbf{m}_{t}, \label{eq:sgd}
\end{align}
with $\mu = 0.9$ and an initial rate $\eta_{0} = 10^{-2}$. Momentum-based descent is preferred over adaptive methods here since the network is trained from random initialization rather than fine-tuned, a regime in which the flatter minima associated with plain momentum are reported to generalize better. The learning rate is halved whenever the training loss fails to improve for five consecutive epochs, with a floor of $10^{-8}$, and optimization stops when the training loss stalls for thirty consecutive epochs, at which point the weights of the best epoch are restored. The maximum budget is $E = 300$ epochs at a batch size of $B = 32$; no run reached the cap, with convergence occurring between epoch 80 and epoch 157 across all configurations. Algorithm~\ref{alg:forward} states the complete forward pass.
\begin{algorithm}[t]
\caption{\nameModel{} forward pass}
\label{alg:forward}
\begin{algorithmic}[1]
\Require Image $\mathbf{X} \in \mathbb{R}^{224 \times 224 \times 3}$; widths $F = (32, 64, 96, 128)$; groups $g = (2, 4, 8, 16)$; rates $p_{s} = (0.05, 0.10, 0.15, 0.20)$
\Ensure Class distribution $\widehat{\mathbf{y}} \in \Delta^{K-1}$
\State $\widehat{\mathbf{X}} \gets \mathcal{A}_{\tau}(\mathbf{X}/255)$ \Comment{Eq.~\eqref{eq:rescale}, \eqref{eq:augment}}
\State $\mathbf{U} \gets \mathrm{MaxPool}_{3,2}(\delta(\mathrm{BN}(\mathbf{W}_{\text{stem}} \circledast_{5,2} \widehat{\mathbf{X}})))$
\For{$l = 1$ \textbf{to} $4$}
  \State $\mathbf{U} \gets \mathrm{ERB}(\mathbf{U}, F_{l}, g_{l}, \text{projection})$ \Comment{Eq.~\eqref{eq:erb1}--\eqref{eq:residual}}
  \State $\mathbf{U} \gets \mathrm{ERB}(\mathbf{U}, F_{l}, g_{l}, \text{identity})$
  \State $\mathbf{U} \gets \mathrm{MSDB}(\mathbf{U}, F_{l}, p_{s,l})$ \Comment{Eq.~\eqref{eq:msdb1}--\eqref{eq:msdb5}}
  \If{$l \leq 2$} \State $\mathbf{U} \gets \mathrm{MaxPool}_{2,2}(\mathbf{U})$ \EndIf
\EndFor
\State $\mathbf{v} \gets [\mathrm{GMP}(\mathbf{U}) ; \mathrm{GAP}(\mathbf{U})]$ \Comment{Eq.~\eqref{eq:dualpool}}
\State \Return $\mathrm{softmax}(\mathbf{W}_{o}^{\top}\mathbf{h}_{2}(\mathbf{v}))$ \Comment{Eq.~\eqref{eq:head1}--\eqref{eq:head3}}
\end{algorithmic}
\end{algorithm}
\subsection{Baselines}
Six ImageNet-pretrained lightweight backbones serve as baselines: EfficientFormerV2-S0~\citep{li2023efficientformerv2}, RepViT-M1.0~\citep{wang2024repvit}, ConvNeXt-Atto~\citep{liu2022convnet}, FastViT-T8~\citep{vasu2023fastvit}, MobileNetV4-Conv-Small~\citep{qin2024mobilenetv4} and GhostNetV2-1.0~\citep{tang2022ghostnetv2}. The selection spans the three design lineages that currently define the sub-five-million parameter regime, namely the purely convolutional stack represented by ConvNeXt-Atto, MobileNetV4-Conv-Small and GhostNetV2-1.0, the reparameterized hybrid represented by RepViT-M1.0 and FastViT-T8, and the efficient-attention transformer represented by EfficientFormerV2-S0, so that the comparison is not confined to a single architectural family. All six are instantiated from the ImageNet-1k checkpoints distributed with the timm library. Every baseline consumes the identical data pipeline, the identical split and the identical resolution, and each is fine-tuned end to end with a freshly initialized classification head using AdamW at a learning rate of $5 \times 10^{-5}$ with weight decay $10^{-3}$, head dropout 0.3 and stochastic depth 0.1, for at most 100 epochs at batch size 32 with early stopping after 15 stagnant epochs and the same halving schedule on the validation loss. The comparison therefore isolates architecture and pretraining, holding data and protocol fixed.
\subsection{Evaluation Protocol}
We report accuracy together with weighted precision, weighted recall and weighted F1, all computed on the held-out test split after model selection on the validation split. Four experiments follow. The seed study repeats the full training procedure at seeds 42 through 46 and reports the mean and the sample standard deviation; the six pretrained baselines are additionally retrained at these same five seeds, and the significance of the accuracy margin is assessed by a two-sided Wilcoxon signed-rank test on the five paired per-seed differences against each baseline, complemented by a paired Student $t$-test, with both families of $p$-values corrected for multiplicity by the Holm step-down procedure~\citep{wilcoxon1945individual, holm1979simple}. The component ablation trains eleven variants that each remove one element of the architecture or of the regularization while holding the seed, the split and the schedule fixed, and reports the resulting accuracy change in percentage points. The cross-dataset study retrains the unmodified architecture on two independently collected datasets, adjusting only the width of the output layer. Interpretability is assessed with Grad-CAM applied to the last convolutional tensor, whose $k$-th activation map is denoted $\mathbf{A}_{k} \in \mathbb{R}^{H' \times W'}$; the importance of that map for class $c$ is the spatially averaged gradient of the class score, and the localization map is the rectified weighted sum of the maps,
\begin{equation}
\alpha_{k}^{c} = \frac{1}{H'W'}\sum_{u}\sum_{w} \frac{\partial \widehat{y}_{c}}{\partial A_{u,w,k}}, \qquad
\mathbf{L}^{c} = \rho\!\left( \sum_{k} \alpha_{k}^{c} \mathbf{A}_{k} \right).
\label{eq:gradcam}
\end{equation}
All experiments were run in TensorFlow 2 with Keras 3.13 on a single NVIDIA P100 accelerator, and the baseline runs additionally used PyTorch with the timm library.

\section{Experimental Results}
\label{sec:results}

\subsection{Classification Performance on \nameDataset}

\begin{figure}[t]
\centering
\includegraphics[width=0.6\linewidth]{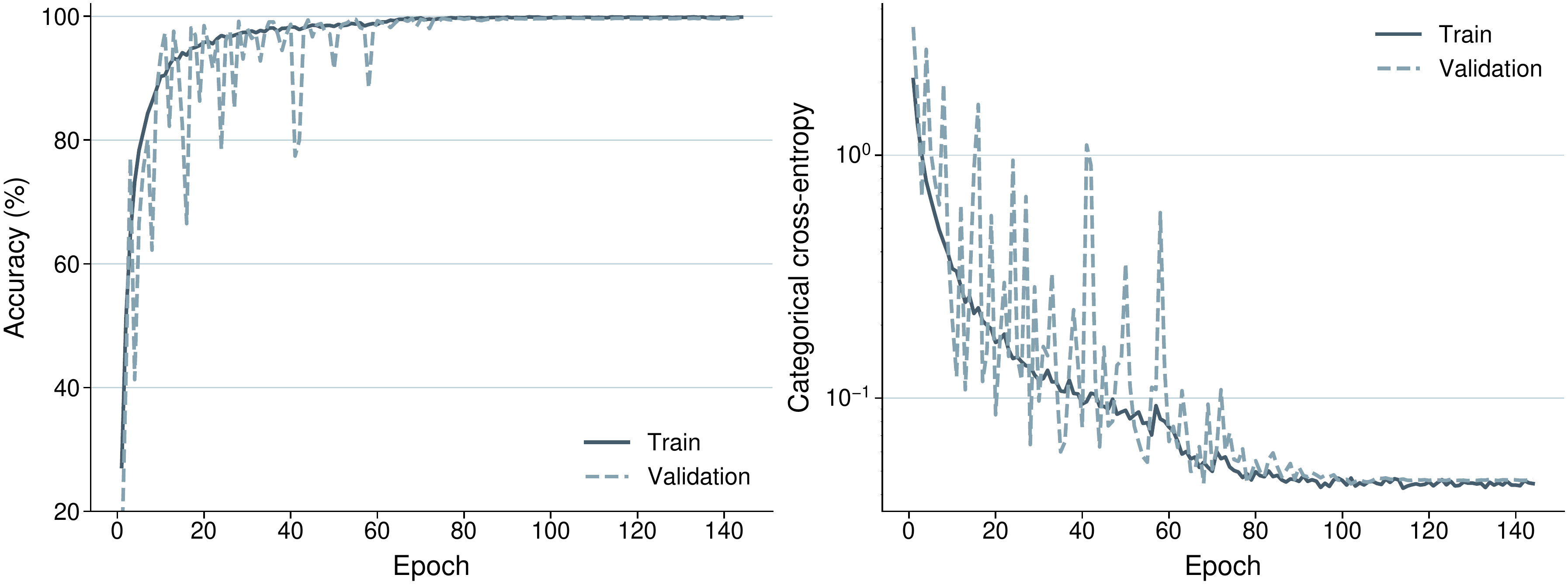}
\caption{Training and validation trajectories of \nameModel{} on \nameDataset. Left: accuracy. Right: categorical cross-entropy on a logarithmic scale.}
\label{fig:training}
\end{figure}

Figure~\ref{fig:training} traces optimization on \nameDataset. Training and validation accuracy separate by less than 0.05 percentage points from epoch 100 onward, and the validation loss tracks the training loss without the upward inflection that marks overfitting, which indicates that the combined effect of spatial dropout, dense dropout, weight decay and embedded augmentation is sufficient at this parameter budget. The best validation epoch is 102 of 145, at which point training accuracy stands at 99.78\% and validation accuracy at 99.73\%.

\begin{figure}[t]
\centering
\includegraphics[width=0.6\linewidth]{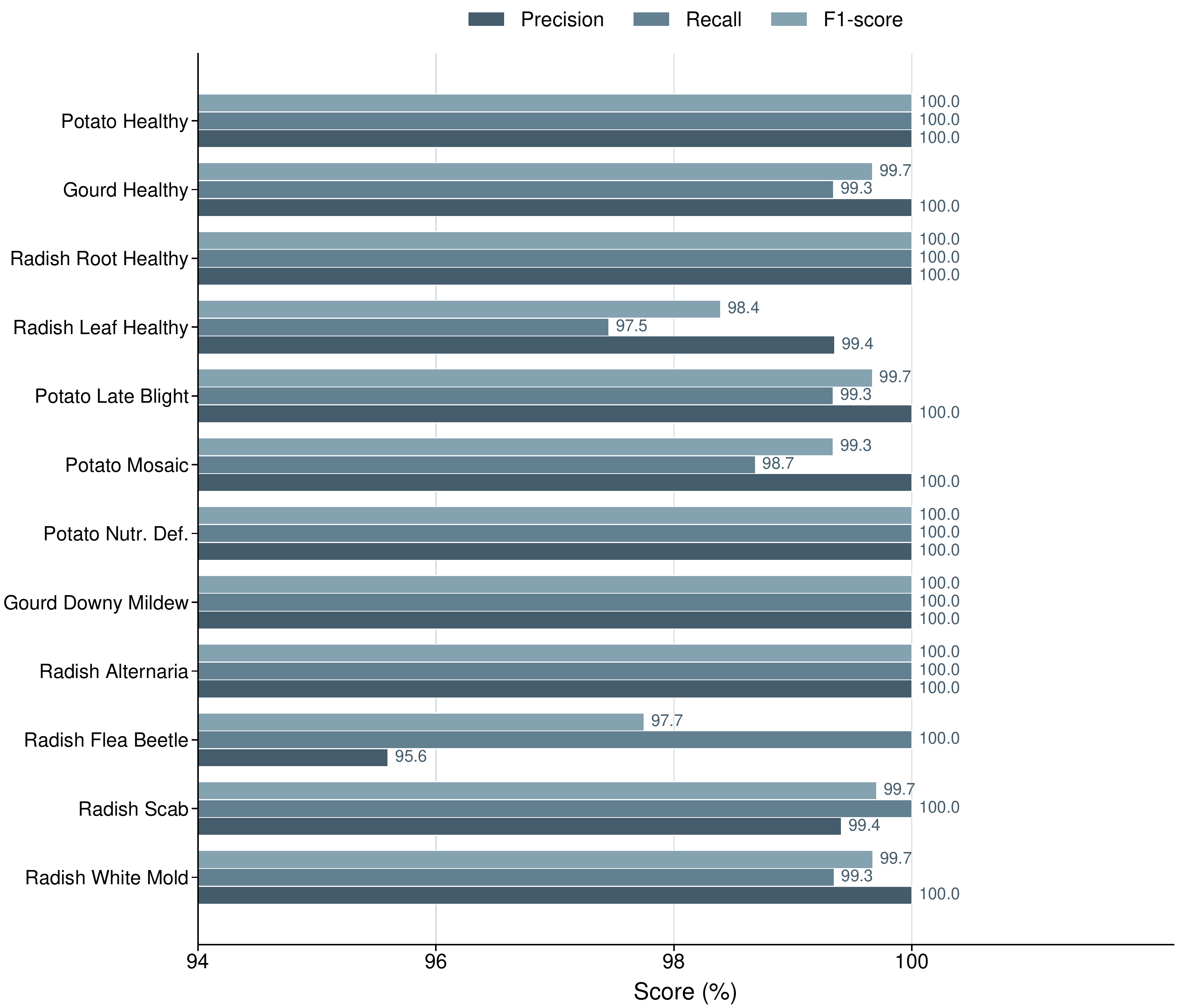}
\caption{Per-class precision, recall and F1 of \nameModel{} on the \nameDataset{} test split.}
\label{fig:per_class}
\end{figure}

Table~\ref{tab:per_class} reports the per-class test performance and Figure~\ref{fig:per_class} presents the same quantities graphically. \nameModel{} attains 99.52\% test accuracy, 99.53\% weighted precision, 99.52\% weighted recall and 99.52\% weighted F1 over the 1,870 held-out images, with a macro F1 of 99.52\% that confirms the aggregate figure is not carried by the larger classes. Five of the twelve classes are classified without error, namely Potato Healthy, Radish Root Healthy, Potato Nutrient Deficiency, Gourd Downy Mildew and Radish \textit{Alternaria brassicae}. The lowest per-class F1 values are 97.75\% for Radish Flea Beetle Damage and 98.39\% for Radish Leaf Healthy, and precision on Radish Flea Beetle Damage falls to 95.60\% while its recall remains at 100\%.

\begin{table}[t]
\centering
\caption{Per-class test performance of \nameModel{} on \nameDataset.}
\label{tab:per_class}
\small
\setlength{\tabcolsep}{4pt}
\begin{tabular}{lrrrr}
\toprule
Class & Prec. (\%) & Rec. (\%) & F1 (\%) & Supp. \\
\midrule
Potato Healthy & 100.00 & 100.00 & 100.00 & 152 \\
Gourd Healthy & 100.00 & 99.34 & 99.67 & 152 \\
Radish Root Healthy & 100.00 & 100.00 & 100.00 & 174 \\
Radish Leaf Healthy & 99.35 & 97.45 & 98.39 & 157 \\
Potato Late Blight & 100.00 & 99.34 & 99.67 & 151 \\
Potato Mosaic & 100.00 & 98.68 & 99.34 & 152 \\
Potato Nutrient Deficiency & 100.00 & 100.00 & 100.00 & 152 \\
Gourd Downy Mildew & 100.00 & 100.00 & 100.00 & 153 \\
Radish \textit{Alternaria brassicae} & 100.00 & 100.00 & 100.00 & 154 \\
Radish Flea Beetle Damage & 95.60 & 100.00 & 97.75 & 152 \\
Radish Scab & 99.41 & 100.00 & 99.70 & 168 \\
Radish White Mold & 100.00 & 99.35 & 99.67 & 153 \\
\midrule
Macro average & 99.53 & 99.51 & 99.52 & 1,870 \\
Weighted average & 99.53 & 99.52 & 99.52 & 1,870 \\
\bottomrule
\end{tabular}
\end{table}

\begin{figure}[t]
\centering
\includegraphics[width=0.6\linewidth]{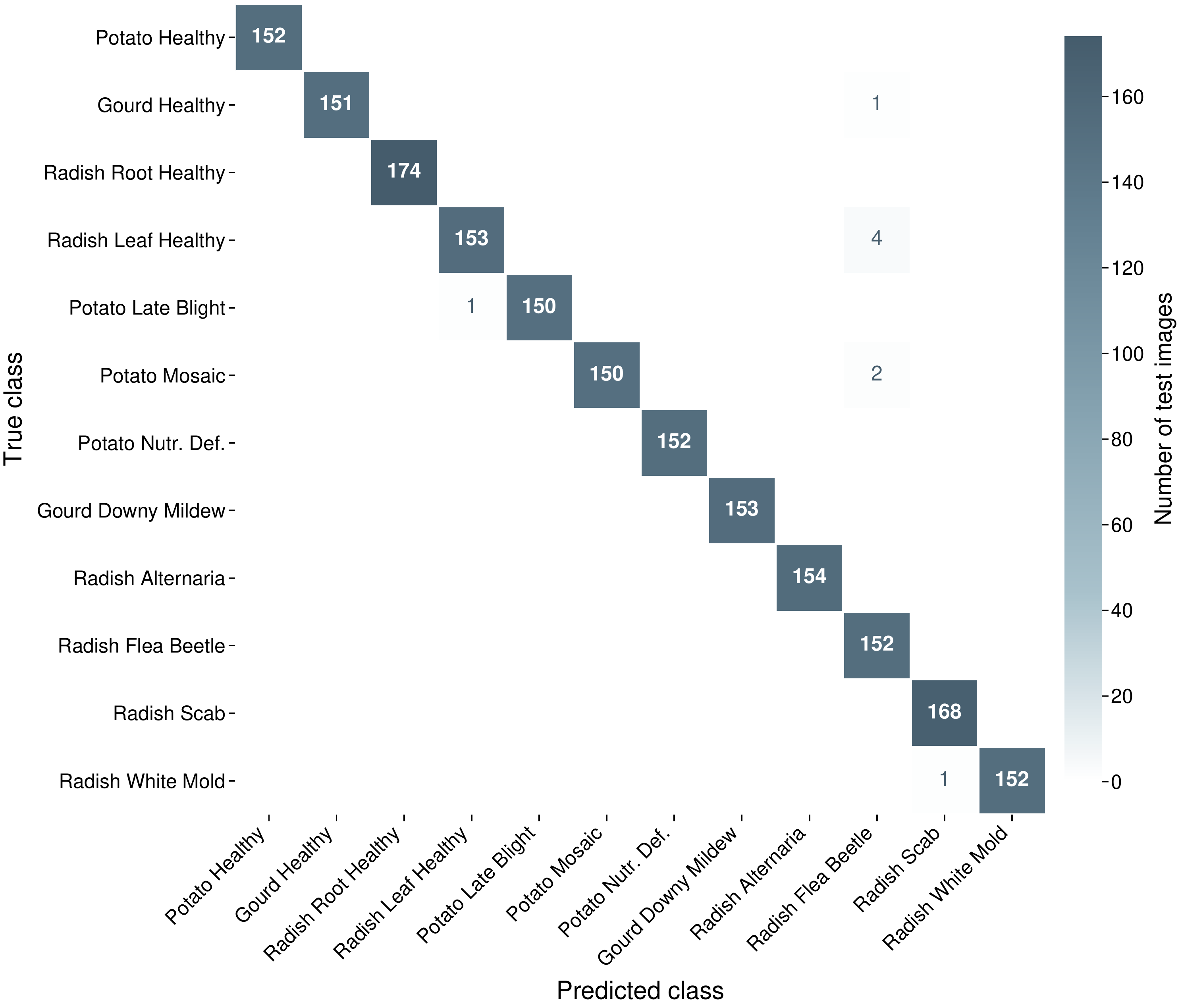}
\caption{Confusion matrix of \nameModel{} on the \nameDataset{} test split. Nine of 1,870 images are misclassified, seven of them into the Radish Flea Beetle Damage class.}
\label{fig:confusion}
\end{figure}

Figure~\ref{fig:confusion} localizes the residual error. Nine of 1,870 test images are misclassified. Seven of the nine fall into the Radish Flea Beetle Damage class, comprising four Radish Leaf Healthy images, two Potato Mosaic images and one Gourd Healthy image; the remaining two are one Potato Late Blight image assigned to Radish Leaf Healthy and one Radish White Mold image assigned to Radish Scab. The dominant confusion is therefore directional and interpretable: flea beetle damage manifests as sparse pinhole punctures of one to two millimeters, and a healthy leaf carrying incidental mechanical damage or a small necrotic fleck presents nearly the same signature at $224 \times 224$ resolution. This failure mode is a resolution limitation rather than a representational one, and higher acquisition resolution over the affected region is the natural remedy.

\subsection{Comparison with Pretrained Lightweight Backbones}

\begin{figure}[t]
\centering
\includegraphics[width=0.4\linewidth]{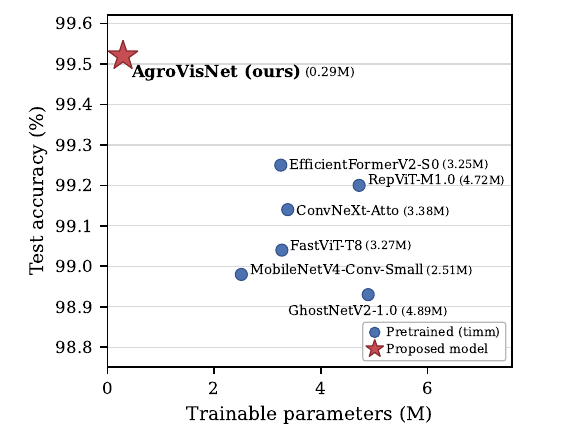}
\caption{Test accuracy against trainable parameter count on \nameDataset.}
\label{fig:params}
\end{figure}

\begin{figure}[t]
\centering
\includegraphics[width=0.6\linewidth]{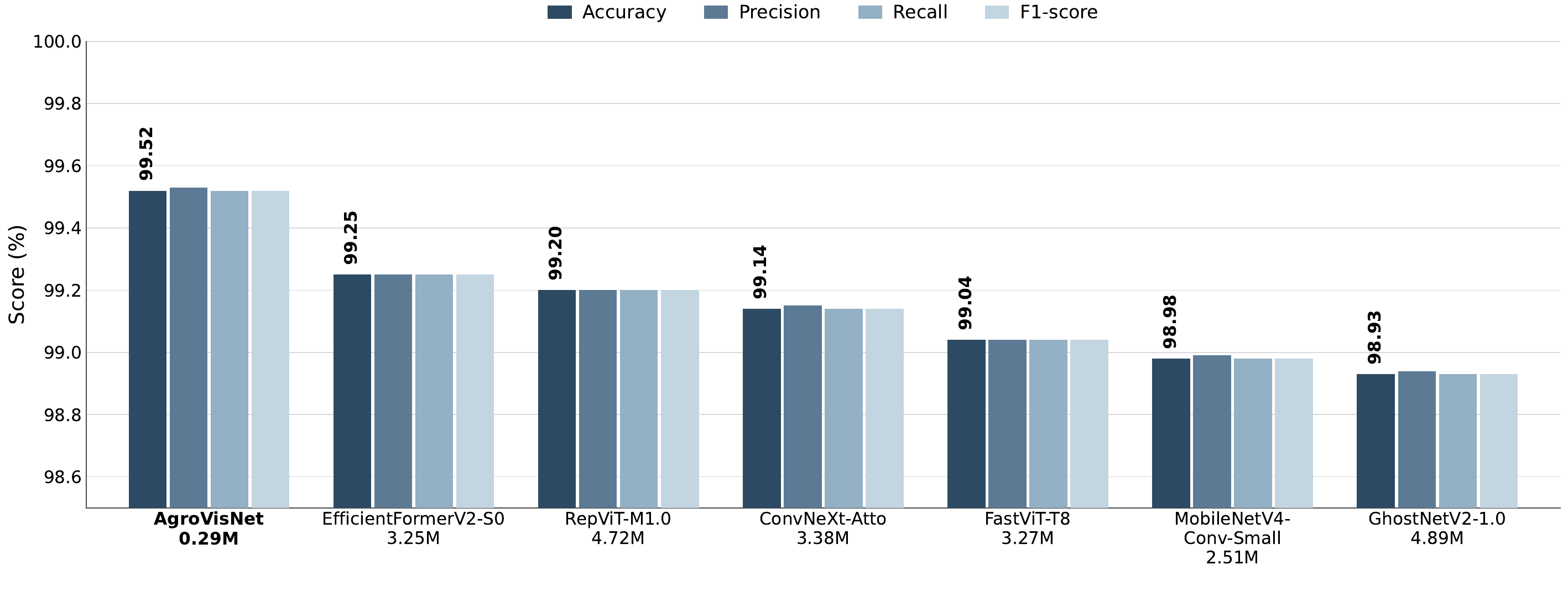}
\caption{Accuracy, precision, recall and F1 of \nameModel{} and the six pretrained baselines on \nameDataset, annotated with the trainable parameter count of each model.}
\label{fig:baselines}
\end{figure}

Table~\ref{tab:baselines} compares \nameModel{} against six ImageNet-pretrained backbones trained under an identical protocol, and Figures~\ref{fig:params} and~\ref{fig:baselines} present the accuracy against the parameter budget and the full metric set respectively. \nameModel{} reaches 99.52\% accuracy at 0.29M trainable parameters. The strongest baseline is EfficientFormerV2-S0 at 99.25\% with 3.25M parameters, followed by RepViT-M1.0 at 99.20\% with 4.72M, ConvNeXt-Atto at 99.14\% with 3.38M, FastViT-T8 at 99.04\% with 3.27M, MobileNetV4-Conv-Small at 98.98\% with 2.51M, and GhostNetV2-1.0 at 98.93\% with 4.89M. The proposed model therefore exceeds the best baseline by 0.27 percentage points while using 11.2 times fewer parameters, and exceeds the smallest baseline by 0.54 percentage points while using 8.7 times fewer parameters; against GhostNetV2-1.0 the parameter ratio reaches 16.8.

\begin{table}[t]
\centering
\caption{\nameModel{} against six ImageNet-pretrained lightweight backbones on \nameDataset, all under an identical split and protocol.}
\label{tab:baselines}
\small
\setlength{\tabcolsep}{4pt}
\begin{tabular}{lrrrrr}
\toprule
Model & Params & Acc. & Prec. & Rec. & F1 \\
 & (M) & (\%) & (\%) & (\%) & (\%) \\
\midrule
\textbf{\nameModel{} (ours)} & \textbf{0.29} & \textbf{99.52} & \textbf{99.53} & \textbf{99.52} & \textbf{99.52} \\
\midrule
EfficientFormerV2-S0   & 3.25 & 99.25 & 99.25 & 99.25 & 99.25 \\
RepViT-M1.0                 & 4.72 & 99.20 & 99.20 & 99.20 & 99.20 \\
ConvNeXt-Atto                   & 3.38 & 99.14 & 99.15 & 99.14 & 99.14 \\
FastViT-T8                     & 3.27 & 99.04 & 99.04 & 99.04 & 99.04 \\
MobileNetV4-Conv-Small       & 2.51 & 98.98 & 98.99 & 98.98 & 98.98 \\
GhostNetV2-1.0 & 4.89 & 98.93 & 98.94 & 98.93 & 98.93 \\
\bottomrule
\end{tabular}
\end{table}

An ensemble of four fully fine-tuned heavyweight backbones provides a further reference point. Averaging the softmax outputs of VGG16, ResNet50, InceptionV3 and MobileNetV2 yields 99.36\% accuracy at a macro-averaged one-versus-rest area under the receiver operating characteristic curve of 0.9973, matching the single strongest lightweight baseline while requiring over 250M parameters and four separate forward passes. \nameModel{} exceeds that ensemble by 0.16 percentage points with a single forward pass through 0.29M parameters, which supports the position that architectural fit to the target distribution matters more than raw capacity on this class of problem.

\subsection{Comparison with Published Plant-Disease Methods}
\label{sec:published}
The pretrained backbones above are general-purpose designs adapted to the task; a complementary question is how \nameModel{} compares against architectures published specifically for plant-disease classification. We reproduce two recent methods: Mob-Res, a MobileNetV2 feature extractor augmented with residual blocks~\citep{pal2025mobres}, and EDL10, a soft-voting ensemble of four convolutional networks trained with the Adam, SGD, RMSprop and Adamax optimizers~\citep{jain2026edl10}. Both were retrained on \nameDataset{} under their own originally published hyperparameters rather than the protocol of Section~\ref{sec:methodology}, so that each method is evaluated with the recipe its authors tuned, and both were assessed on the same 1{,}870-image test split (Table~\ref{tab:published}). EDL10's reported parameter count is the total across its four sub-models, and the \nameModel{} accuracy and F1 figures repeat Table~\ref{tab:baselines}.

Mob-Res reaches 99.20\% accuracy and F1 at 3.47M parameters and EDL10 reaches 96.52\% accuracy at 8.49M parameters across its four sub-models, against \nameModel{}'s 99.52\% at 0.29M. The proposed model is therefore 0.32 and 3.00 percentage points more accurate while using 12 and 29 times fewer parameters than the two published designs, and the ensemble in particular incurs four forward passes for the lowest accuracy of the three. The headline figure each method reports on its own dataset is listed for context but is not comparable across rows, because the datasets and class inventories differ; the informative comparison is the reproduced column, measured on a common dataset and test split, which reinforces the pattern established against the pretrained backbones, that a compact architecture matched to the target distribution is more accurate and far smaller than heavier general-purpose or ensemble alternatives.

\begin{table}[t]
\centering
\caption{Comparison of \nameModel{} against two methods published specifically for plant-disease classification, each retrained on \nameDataset{} under its own published hyperparameters and evaluated on the shared 1,870-image test split (Section~\ref{sec:published}).}
\label{tab:published}
\small
\setlength{\tabcolsep}{4pt}
\begin{tabular}{lrrrr}
\toprule
Method & Params & Orig.\ acc. & Acc. & F1 \\
 &  & (own data) & (\%) & (\%) \\
\midrule
Mob-Res~\citep{pal2025mobres}   & 3.47M & 99.47 & 99.20 & 99.20 \\
EDL10~\citep{jain2026edl10}     & 8.49M & 97.00 & 96.52 & 96.53 \\
\nameModel{} (proposed)         & 0.29M & ---   & 99.52 & 99.52 \\
\bottomrule
\end{tabular}
\end{table}

\subsection{Deployment and Computational Efficiency}
\label{sec:quantization}
Parameter count alone does not determine whether a model is deployable; the arithmetic executed per image and the memory the model occupies matter as much on the low-cost hardware that is the intended deployment target. Table~\ref{tab:efficiency} therefore reports the multiply--accumulate (MAC) cost, the single-precision on-disk size and the test accuracy of \nameModel{} against the six pretrained backbones at $224 \times 224$ resolution. MAC counts are measured directly, single-precision size follows the parameter count at four bytes per weight, and the parameter and accuracy columns repeat Table~\ref{tab:baselines}; of \nameModel's 296,780 total parameters, 290,572 are trainable. The final row of Table~\ref{tab:efficiency} reports the baseline-to-\nameModel ratio from the least to the most demanding baseline on each axis, so that a larger value favours \nameModel. Because wall-clock latency is a property of the measurement platform rather than of the architecture, the MAC count serves as the hardware-independent compute proxy across models, and measured latency is reported for \nameModel{} alone.

\begin{table}[t]
\centering
\caption{Computational efficiency of \nameModel{} against the six pretrained backbones on \nameDataset{} at $224 \times 224$ resolution (Section~\ref{sec:quantization}).}
\label{tab:efficiency}
\small
\setlength{\tabcolsep}{3pt}
\begin{tabular}{lrrrr}
\toprule
Model & Params & MACs & FP32 & Acc. \\
 & (M) & (M) & (MB) & (\%) \\
\midrule
\textbf{\nameModel{} (ours)} & \textbf{0.29} & \textbf{132.70} & \textbf{1.13} & \textbf{99.52} \\
\midrule
EfficientFormerV2-S0   & 3.25 & 406.70  & 12.43 & 99.25 \\
RepViT-M1.0            & 4.72 & 1124.58 & 24.56 & 99.20 \\
ConvNeXt-Atto          & 3.38 & 551.75  & 12.92 & 99.14 \\
FastViT-T8             & 3.27 & 539.11  & 12.54 & 99.04 \\
MobileNetV4-Conv-Small & 2.51 & 188.75  & 9.70  & 98.98 \\
GhostNetV2-1.0         & 4.89 & 176.34  & 18.79 & 98.93 \\
\midrule
\textit{ratio (min--max)} & \textit{8.7--16.8$\times$} & \textit{1.3--8.5$\times$} & \textit{8.6--21.7$\times$} & --- \\
\bottomrule
\end{tabular}
\end{table}

At 132.70M MACs, \nameModel{} is the cheapest model in the comparison, from 1.3 times below the next-lightest baseline, GhostNetV2-1.0, to 8.5 times below the heaviest, RepViT-M1.0; its single forward pass costs 265.40M floating-point operations, twice the MAC count. It is also the smallest on disk at 1.13 MB, against 9.70 to 24.56 MB for the baselines, a direct consequence of a parameter budget that is 8.7 to 16.8 times below the 2.51 to 4.89M parameters of the pretrained backbones. \nameModel{} is thus at once the most accurate model in the comparison and the least expensive to evaluate, so that on this dataset no competing model is simultaneously more accurate and cheaper to run.

\begin{table}[t]
\centering
\caption{Export and post-training quantization of \nameModel{}, measured through the TensorFlow Lite XNNPACK delegate (Section~\ref{sec:quantization}).}
\label{tab:quantization}
\small
\setlength{\tabcolsep}{4pt}
\begin{tabular}{lrrr}
\toprule
Export format & Size & CPU lat. & Acc. \\
 & (MB) & (ms) & (\%) \\
\midrule
Keras FP32           & 1.13 & ---            & 99.52 \\
TFLite FP32          & 1.17 & 10.44          & 99.52 \\
TFLite dynamic-range & 0.43 & 15.12          & 99.52 \\
TFLite INT8 (full)   & 0.46 & \phantom{0}8.40 & 99.30 \\
\bottomrule
\end{tabular}
\end{table}

\begin{figure}[t]
\centering
\includegraphics[width=0.6\linewidth]{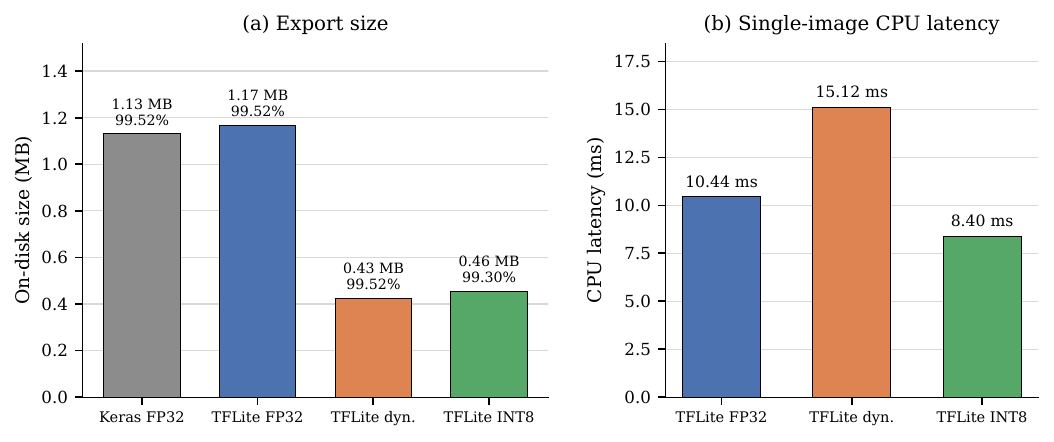}
\caption{Export and quantization behaviour of \nameModel{}. (a) On-disk size across export formats, annotated with test accuracy. (b) Single-image CPU latency of the TensorFlow Lite formats through the XNNPACK delegate.}
\label{fig:quantization}
\end{figure}

Table~\ref{tab:quantization} and Figure~\ref{fig:quantization} report how the model behaves once exported for on-device use. The single-precision TensorFlow Lite export reproduces the Keras accuracy exactly at 1.17 MB; dynamic-range quantization preserves accuracy to the fourth decimal while reducing the model to 0.43 MB; and full-integer INT8 quantization retains accuracy to within 0.22 percentage points at 0.46 MB, small enough to sit in the flash budget of a low-cost microcontroller. The INT8 model is additionally the fastest of the three at 8.40 ms per image, below both the 10.44 ms of the single-precision export and the 9.00 ms single-image latency measured on a GPU, because integer kernels map cleanly onto the vector units of a commodity CPU; the dynamic-range variant is the slowest at 15.12 ms, as its weights are dequantised on the fly. Taken together, these measurements turn the parameter budget into a concrete deployment envelope: a sub-half-megabyte model that classifies an image in under ten milliseconds on a CPU with no dedicated accelerator.

\begin{figure}[t]
\centering
\includegraphics[width=0.6\linewidth]{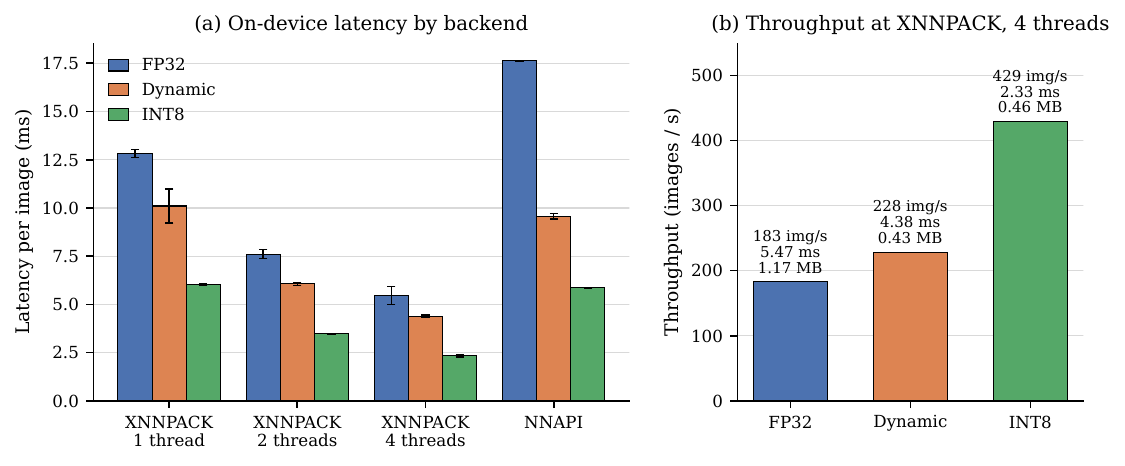}
\caption{On-device deployment of \nameModel{} on the Android handset of Table~\ref{tab:edge}. (a) Single-image latency across the XNNPACK CPU delegate and NNAPI, grouped by export format; error bars are the standard deviation across three repeats. (b) Throughput of the headline four-thread XNNPACK configuration.}
\label{fig:edge}
\end{figure}

Compute cost and quantized size predict deployability; a direct measurement confirms it. We exported \nameModel{} to TensorFlow Lite and benchmarked all three formats on a mid-range Android handset (realme RMX3853, Qualcomm Snapdragon 7+ Gen~3, 12\,GB RAM, Android~16) with the \texttt{benchmark\_model} tool pinned to the five performance cores, over 100 runs per measurement and three repeats. Each configuration was run for 100 measurements after 20 warmup runs, with the standard deviation across the three repeats remaining below 0.9 ms in every case; the GPU delegate was also requested but fell back to CPU execution on this device and is therefore omitted from the comparison. Table~\ref{tab:edge} and Figure~\ref{fig:edge} report single-image latency across the XNNPACK CPU delegate at one, two and four threads and the Android Neural Networks API (NNAPI). The full-integer INT8 model is the fastest configuration at every thread count, reaching 2.33\,ms per image at four threads, approximately 429 images per second, against 4.38\,ms for the dynamic-range model and 5.47\,ms for the single-precision export. Latency scales cleanly with the thread count, the INT8 model falling from 6.02\,ms at one thread to 3.47\,ms at two and 2.33\,ms at four, a 2.6-fold reduction across the four-core span, while NNAPI improves on none of the four-thread XNNPACK paths and reads 5.85\,ms for INT8 as the delegate falls back to CPU execution, so the CPU delegate is the operative deployment path on this system-on-chip. The 2.33\,ms on-device INT8 latency is below the 8.40\,ms desktop-CPU figure of Table~\ref{tab:quantization}, because integer kernels map directly onto the mobile vector units, and it places recognition far above video frame rate on a handset with no dedicated accelerator.
Together, these measurements substantiate the deployability claim in the paper's title: the model is not merely small in parameter count but cheap to compute, sub-megabyte on disk, and real-time on a commodity smartphone, matching or exceeding the accuracy of backbones nearly an order of magnitude larger on every efficiency axis reported above.

\begin{table}[t]
\centering
\caption{On-device inference of \nameModel{} on a mid-range Android handset (realme RMX3853, Snapdragon 7+ Gen 3, Android 16), measured with the TensorFlow Lite \texttt{benchmark\_model} tool (Section~\ref{sec:quantization}). Throughput is for the four-thread XNNPACK configuration.}
\label{tab:edge}
\small
\setlength{\tabcolsep}{3pt}
\begin{tabular}{lrrrrrr}
\toprule
Export format & Size & \multicolumn{3}{c}{XNNPACK (ms)} & NNAPI & Thr. \\
\cmidrule(lr){3-5}
 & (MB) & 1\,t & 2\,t & 4\,t & (ms) & (img/s) \\
\midrule
TFLite FP32          & 1.17 & 12.83 & 7.61 & 5.47 & 17.64 & 182.9 \\
TFLite dynamic-range & 0.43 & 10.10 & 6.06 & 4.38 & \phantom{0}9.57 & 228.4 \\
TFLite INT8 (full)   & 0.46 & \phantom{0}6.02 & 3.47 & 2.33 & \phantom{0}5.85 & 429.1 \\
\bottomrule
\end{tabular}
\end{table}

\subsection{Stability Across Random Seeds}

\begin{figure}[t]
\centering
\includegraphics[width=0.6\linewidth]{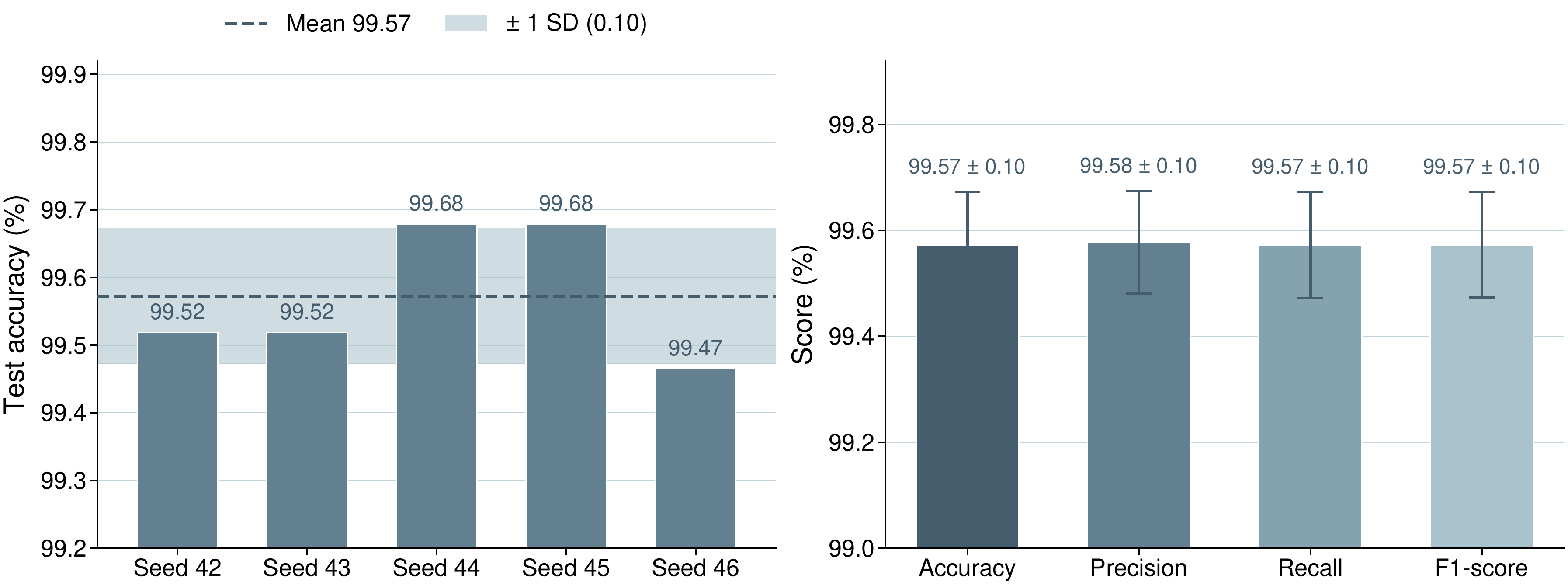}
\caption{Seed-to-seed stability of \nameModel{} on \nameDataset. Left: test accuracy per seed against the mean and the one-standard-deviation band. Right: mean and standard deviation of the four aggregate metrics.}
\label{fig:multiseed}
\end{figure}

\begin{figure}[t]
\centering
\includegraphics[width=0.6\linewidth]{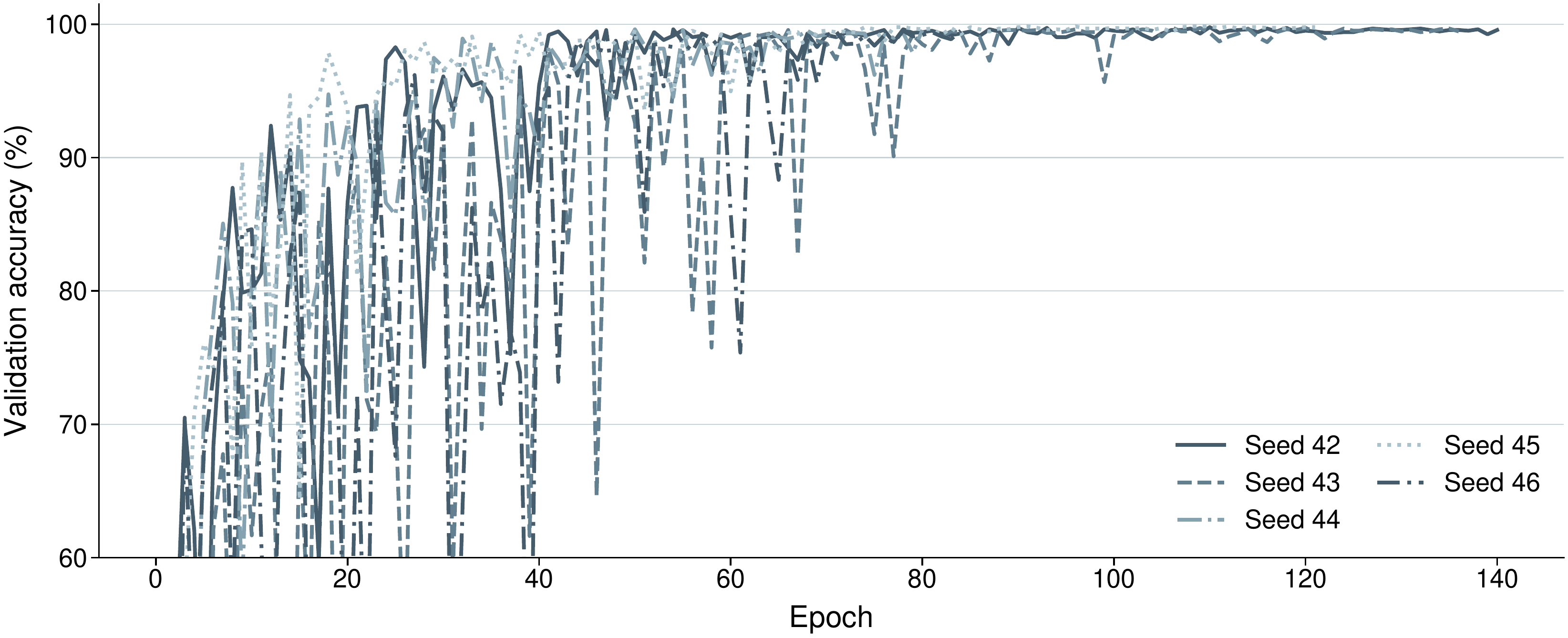}
\caption{Validation accuracy trajectories of the five seeds. All runs exceed 99\% by epoch 60 despite differing convergence behavior over the first 40 epochs.}
\label{fig:multiseed_curves}
\end{figure}

Table~\ref{tab:multiseed} and Figure~\ref{fig:multiseed} report five independent repetitions of the full training procedure at seeds 42 through 46. Test accuracy averages 99.57\% with a sample standard deviation of 0.10 percentage points and a range of 99.47\% to 99.68\%; weighted F1 tracks it at $99.57 \pm 0.10$\%. The lower bound of the observed range exceeds the accuracy of every pretrained baseline, which indicates that the margin reported in Table~\ref{tab:baselines} is not an artifact of a favorable initialization. Convergence is less stable than accuracy: the best epoch varies from 50 to 110 and training time from 56.6 to 98.7 minutes, a spread driven by the training-loss-triggered learning rate schedule, which reaches its final rate after two halvings in some runs and four in others. Figure~\ref{fig:multiseed_curves} shows the corresponding validation trajectories, all of which exceed 99\% by epoch 60 despite differing sharply during the first 40 epochs.

\begin{table}[t]
\centering
\caption{Five-seed repetition of the full training procedure on \nameDataset.}
\label{tab:multiseed}
\small
\setlength{\tabcolsep}{4pt}
\begin{tabular}{rrrrrr}
\toprule
Seed & Epochs & Best & Val. acc. & Test acc. & F1 \\
 & run & epoch & (\%) & (\%) & (\%) \\
\midrule
42 & 140 & 110 & 99.79 & 99.52 & 99.52 \\
43 & 136 & 106 & 99.79 & 99.52 & 99.52 \\
44 & 80 & 50 & 99.63 & 99.68 & 99.68 \\
45 & 121 & 91 & 99.84 & 99.68 & 99.68 \\
46 & 84 & 54 & 99.63 & 99.47 & 99.47 \\
\midrule
\multicolumn{3}{l}{Mean $\pm$ SD} & 99.74 $\pm$ 0.10 & 99.57 $\pm$ 0.10 & 99.57 $\pm$ 0.10 \\
\bottomrule
\end{tabular}
\end{table}

\subsection{Statistical Significance of the Margin over the Baselines}
\label{sec:significance}

\begin{figure}[t]
\centering
\includegraphics[width=0.6\linewidth]{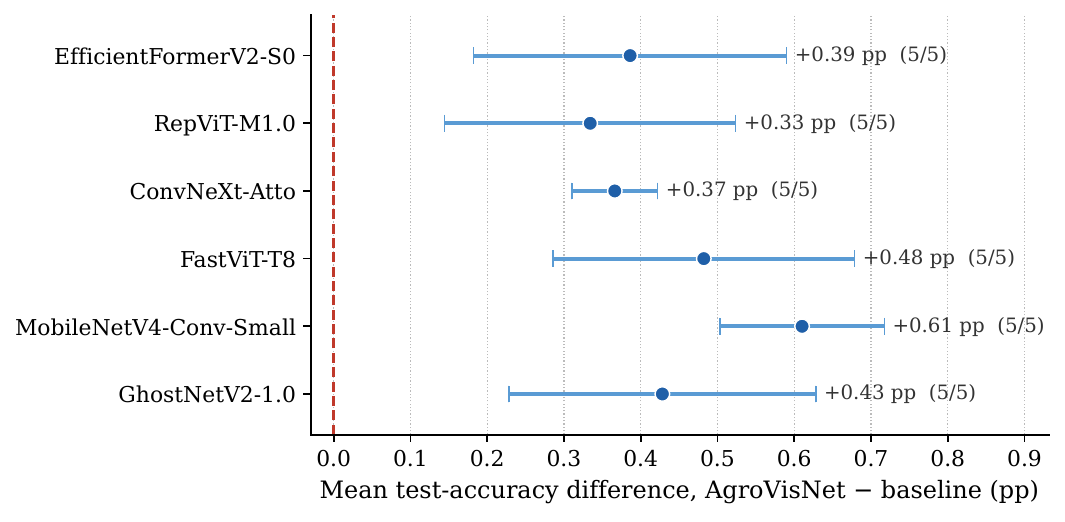}
\caption{Paired per-seed test-accuracy difference between \nameModel{} and each pretrained backbone on \nameDataset{}, over the five seeds of Table~\ref{tab:multiseed}. Each marker is the mean difference (\nameModel{} minus baseline) with its 95\% CI whisker; the dashed line marks equality.}
\label{fig:significance}
\end{figure}

A margin of a few tenths of a percentage point invites the question of whether it reflects a genuine difference or a favourable draw of the single reference seed. To settle this, each of the six pretrained backbones was retrained under its own protocol at the same five seeds used for the seed study of Table~\ref{tab:multiseed}, namely 42 through 46, and the test accuracy of \nameModel{} was paired with that of each baseline seed by seed. The five paired differences per baseline were tested with a two-sided Wilcoxon signed-rank test, the nonparametric procedure standardly recommended for classifier comparison, and the six resulting $p$-values were corrected for multiplicity with the Holm step-down procedure~\citep{wilcoxon1945individual, holm1979simple, demsar2006statistical}. A paired two-sided Student $t$-test, Holm-corrected across the same six comparisons, is reported alongside as a parametric complement. Table~\ref{tab:significance} and Figure~\ref{fig:significance} report the outcome.

\begin{table}[t]
\centering
\caption{Paired per-seed significance of the \nameModel{} accuracy margin over the six pretrained backbones (Table~\ref{tab:multiseed} seeds). $\overline{\Delta}$: mean accuracy difference with 95\% CI; W/L: win/loss record; $p_{\mathrm{H}}^{W}$, $p_{\mathrm{H}}^{t}$: Holm-adjusted Wilcoxon and paired $t$-test $p$-values (Section~\ref{sec:significance}).}
\label{tab:significance}
\small
\setlength{\tabcolsep}{2.1pt}
\begin{tabular}{lccccc}
\toprule
Baseline & $\overline{\Delta}$ (pp) & 95\% CI (pp) & W/L & $p_{\mathrm{H}}^{W}$ & $p_{\mathrm{H}}^{t}$ \\
\midrule
EfficientFormerV2-S0   & $+0.39$ & $[0.18,\,0.59]$ & 5/0 & 0.375 & 0.013 \\
RepViT-M1.0            & $+0.33$ & $[0.14,\,0.52]$ & 5/0 & 0.375 & 0.013 \\
ConvNeXt-Atto          & $+0.37$ & $[0.31,\,0.42]$ & 5/0 & 0.375 & $<0.001$ \\
FastViT-T8             & $+0.48$ & $[0.29,\,0.68]$ & 5/0 & 0.375 & 0.010 \\
MobileNetV4-Conv-Small & $+0.61$ & $[0.50,\,0.72]$ & 5/0 & 0.375 & $<0.001$ \\
GhostNetV2-1.0         & $+0.43$ & $[0.23,\,0.63]$ & 5/0 & 0.375 & 0.012 \\
\bottomrule
\end{tabular}
\end{table}

\nameModel{} is more accurate than every baseline on all five of the five seeds, with no ties, and the mean paired difference ranges from 0.33 percentage points against RepViT-M1.0 to 0.61 against MobileNetV4-Conv-Small. The 95\% confidence interval on that difference excludes zero for all six baselines, the tightest being $[0.31, 0.42]$ against ConvNeXt-Atto and the widest $[0.14, 0.52]$ against RepViT-M1.0 (Figure~\ref{fig:significance}). The paired $t$-test rejects the null of equal accuracy for all six after Holm correction, with adjusted $p$-values from $3.2 \times 10^{-4}$ against ConvNeXt-Atto to 0.013 against EfficientFormerV2-S0 and RepViT-M1.0, so the margin is significant at the 0.05 level against every backbone.

The Wilcoxon signed-rank test, by contrast, reaches significance against none of the six, and the reason is a property of the sample size rather than of the effect. With five nonzero paired differences that all share a sign, the exact two-sided Wilcoxon $p$-value attains its smallest possible value, $2/2^{5} = 0.0625$, which already exceeds 0.05 before any correction; every baseline therefore records a raw $p$ of exactly 0.0625, which Holm inflates to 0.375. The rank test is thus at its resolution floor and cannot certify a difference at five seeds no matter how large or how consistent it is. The convergent evidence of a clean five-of-five sweep, six confidence intervals that exclude zero and a significant paired $t$-test establishes that the accuracy advantage of \nameModel{} over the pretrained backbones is reliable rather than an artifact of initialization; a larger seed budget is the only requirement for the rank test to reach the same conclusion.

\subsection{Component Ablation}

\begin{figure}[t]
\centering
\includegraphics[width=0.6\linewidth]{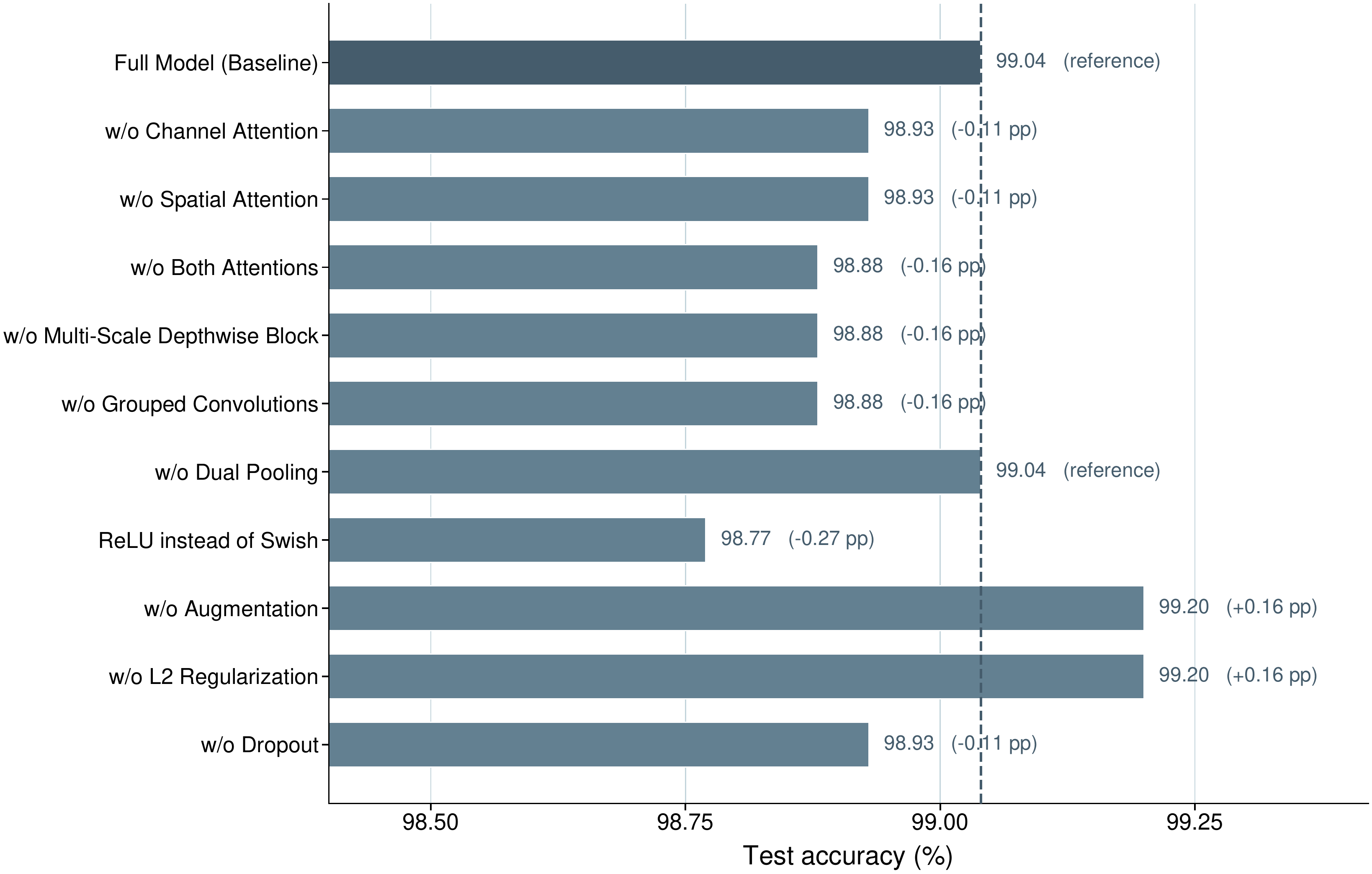}
\caption{Test accuracy of ten ablation variants and the full-model baseline. The dashed line marks the full model and each bar is annotated with its change in percentage points.}
\label{fig:ablation}
\end{figure}

\begin{figure}[t]
\centering
\includegraphics[width=0.6\linewidth]{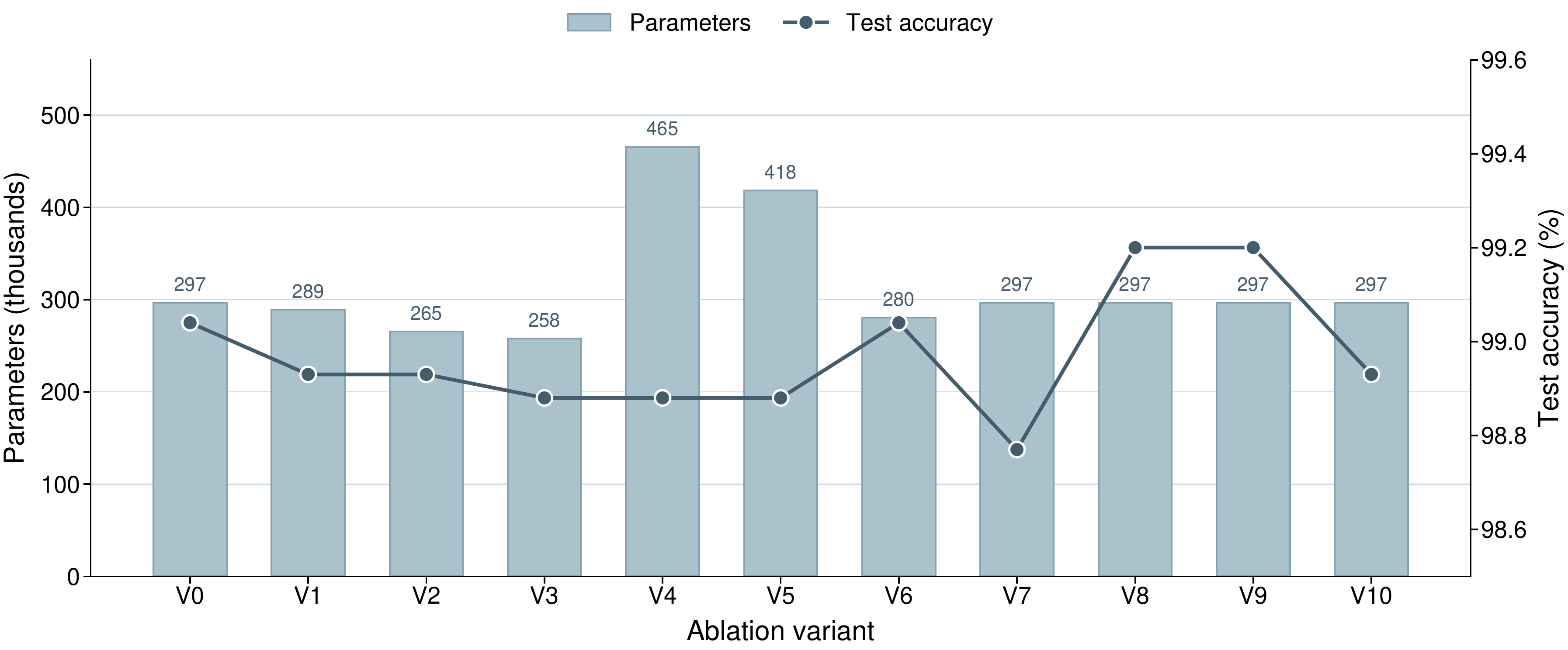}
\caption{Parameter count and test accuracy of each ablation variant. Removing the multi-scale depthwise block (V4) or the grouped convolutions (V5) increases the parameter count while lowering accuracy.}
\label{fig:ablation_params}
\end{figure}

Table~\ref{tab:ablation} and Figures~\ref{fig:ablation} and~\ref{fig:ablation_params} report ten single-component variants against the full-model baseline while holding the seed, the split and the schedule fixed. The ablation protocol re-seeds the entire stack before every variant and follows a single fixed trajectory, under which the full model reaches 99.04\% rather than the 99.52\% of the main run; all comparisons below are therefore made within the ablation protocol.

Substituting the rectified linear unit for swish is the single most damaging change at $-0.27$ percentage points, which is consistent with the smoothness of swish being useful at this depth. Removing both attention gates, the multi-scale depthwise block or the grouped convolutions each costs $-0.16$ percentage points, and removing either gate individually costs $-0.11$. The parameter accounting in Figure~\ref{fig:ablation_params} is the more consequential reading. Removing the Multi-Scale Depthwise Block raises the parameter count from 296,780 to 465,420, an increase of 56.8\%, while lowering accuracy; removing the grouped convolutions raises it to 418,316, an increase of 41.0\%, again while lowering accuracy. The two mechanisms therefore contribute accuracy and parameter economy jointly rather than trading one against the other, which is the central design claim of the architecture. The two attention gates together cost 39,040 parameters, which is 13.2\% of the total, for a combined 0.16 percentage point contribution.

Two variants improve on the full model within this protocol. Removing augmentation and removing weight decay each yield 99.20\%, an increase of 0.16 percentage points, and removing augmentation additionally cuts training time from 108.5 to 32.2 minutes. We report this without adjustment. \nameDataset{} is class balanced and was acquired under a controlled illumination protocol, and on a test split drawn from that same acquisition process the regularizers restrict the fit at a small cost in clean accuracy. Both are retained in the released configuration since the deployment target is field imagery under illumination, viewpoint and background conditions absent from the current test split, and the field-condition evaluation listed among the outstanding items is the experiment that will settle this trade-off quantitatively.

\begin{table}[t]
\centering
\caption{Component ablation on \nameDataset. Accuracy change is measured against the V0 full model within the ablation protocol.}
\label{tab:ablation}
\small
\setlength{\tabcolsep}{3.5pt}
\begin{tabular}{llrrr}
\toprule
ID & Removed component & Params & Acc. (\%) & $\Delta$ (pp) \\
\midrule
V0 & Full model (baseline) & 296,780 & 99.04 & reference \\
V1 & Channel attention & 289,100 & 98.93 & $-0.11$ \\
V2 & Spatial attention & 265,420 & 98.93 & $-0.11$ \\
V3 & Both attentions & 257,740 & 98.88 & $-0.16$ \\
V4 & Multi-Scale Depthwise Block & 465,420 & 98.88 & $-0.16$ \\
V5 & Grouped convolutions & 418,316 & 98.88 & $-0.16$ \\
V6 & Dual pooling & 280,396 & 99.04 & $0.00$ \\
V7 & Swish, replaced by ReLU & 296,780 & 98.77 & $-0.27$ \\
V8 & Augmentation & 296,780 & 99.20 & $+0.16$ \\
V9 & $\ell_{2}$ regularization & 296,780 & 99.20 & $+0.16$ \\
V10 & Dropout & 296,780 & 98.93 & $-0.11$ \\
\bottomrule
\end{tabular}
\end{table}

\subsection{Cross-Dataset Generalization}

\begin{figure}[t]
\centering
\includegraphics[width=0.6\linewidth]{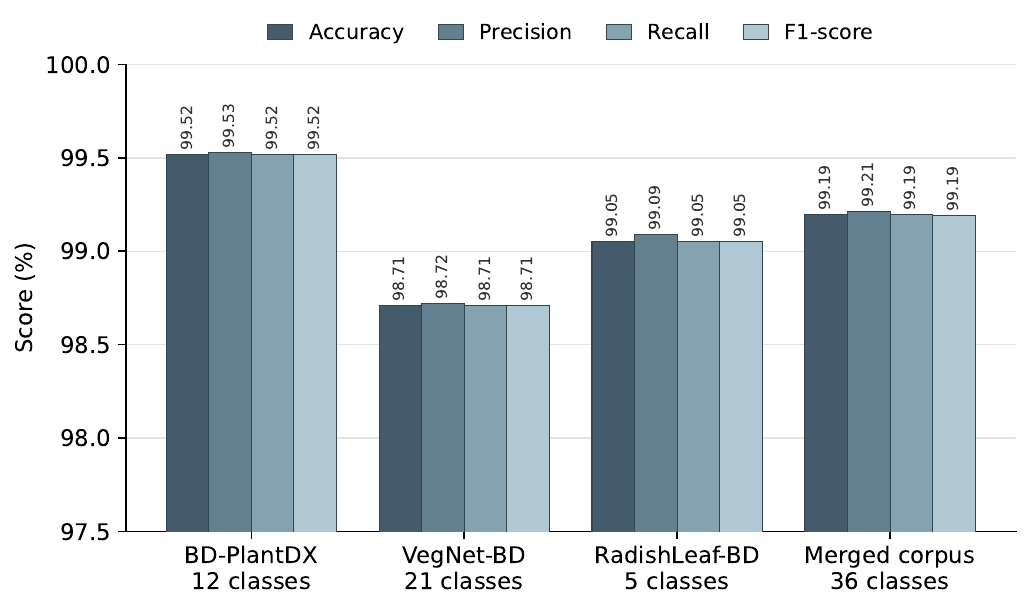}
\caption{Accuracy, precision, recall and F1 of the unmodified \nameModel{} architecture across \nameDataset, VegNet-BD, RadishLeaf-BD and the merged 36-class corpus of all three.}
\label{fig:cross}
\end{figure}

Table~\ref{tab:cross} and Figure~\ref{fig:cross} report the unmodified architecture retrained on two independently collected Bangladeshi datasets, with the output width as the only change. On VegNet-BD, comprising 12,786 images across 21 classes of six vegetable crops~\citep{hasan2024comprehensive}, \nameModel{} attains 98.71\% accuracy and 98.71\% weighted F1 at 291,157 parameters. On RadishLeaf-BD, comprising 2,801 images across five radish classes~\citep{hasan2025smartphone}, it attains 99.05\% accuracy and 99.05\% weighted F1 at 290,117 parameters. Neither transfer required a change to the block structure, the widths, the group counts, the augmentation or the optimizer, which addresses the concern that a small architecture tuned on one collection encodes a dataset-specific inductive bias.

\begin{table}[t]
\centering
\caption{Generalization of the unmodified \nameModel{} architecture to two independently collected datasets and to the merged 36-class corpus of all three.}
\label{tab:cross}
\small
\setlength{\tabcolsep}{2.5pt}
\begin{tabular}{lrrrrr}
\toprule
Dataset & Cls. & Images & Params & Acc. (\%) & F1 (\%) \\
\midrule
\nameDataset{} (ours) & 12 & 12,432 & 290,572 & 99.52 & 99.52 \\
VegNet-BD & 21 & 12,786 & 291,157 & 98.71 & 98.71 \\
RadishLeaf-BD & 5 & 2,801 & 290,117 & 99.05 & 99.05 \\
\midrule
Merged corpus (3 datasets) & 36 & 28,019 & 292,132 & 99.19 & 99.19 \\
\bottomrule
\end{tabular}
\end{table}

\begin{figure}[t]
\centering
\includegraphics[width=0.6\linewidth]{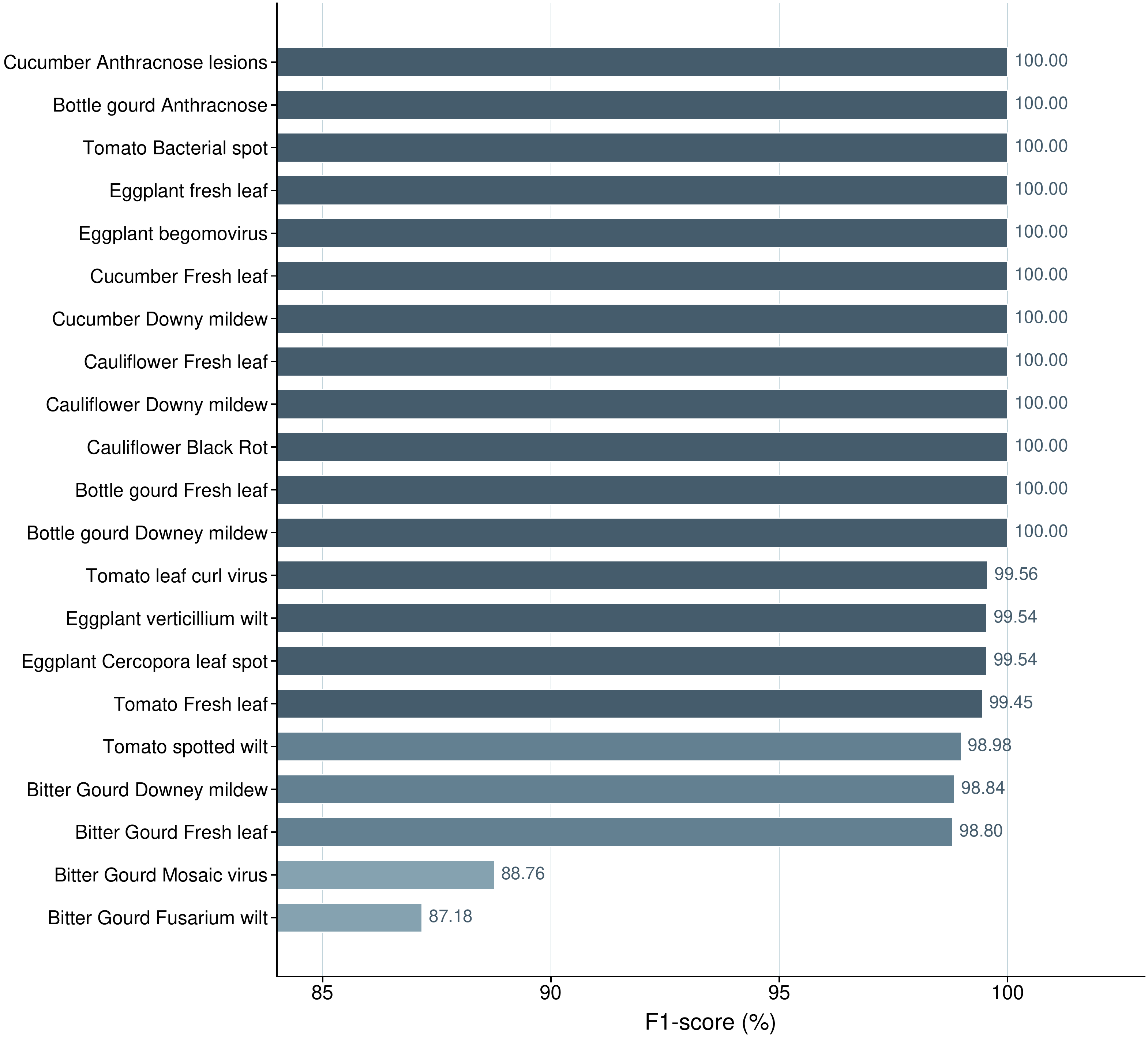}
\caption{Per-class F1 of \nameModel{} on the 21 classes of VegNet-BD. Twelve classes are classified without error and the aggregate is limited by four bitter gourd classes.}
\label{fig:vegnet}
\end{figure}

\begin{figure}[t]
\centering
\includegraphics[width=0.4\linewidth]{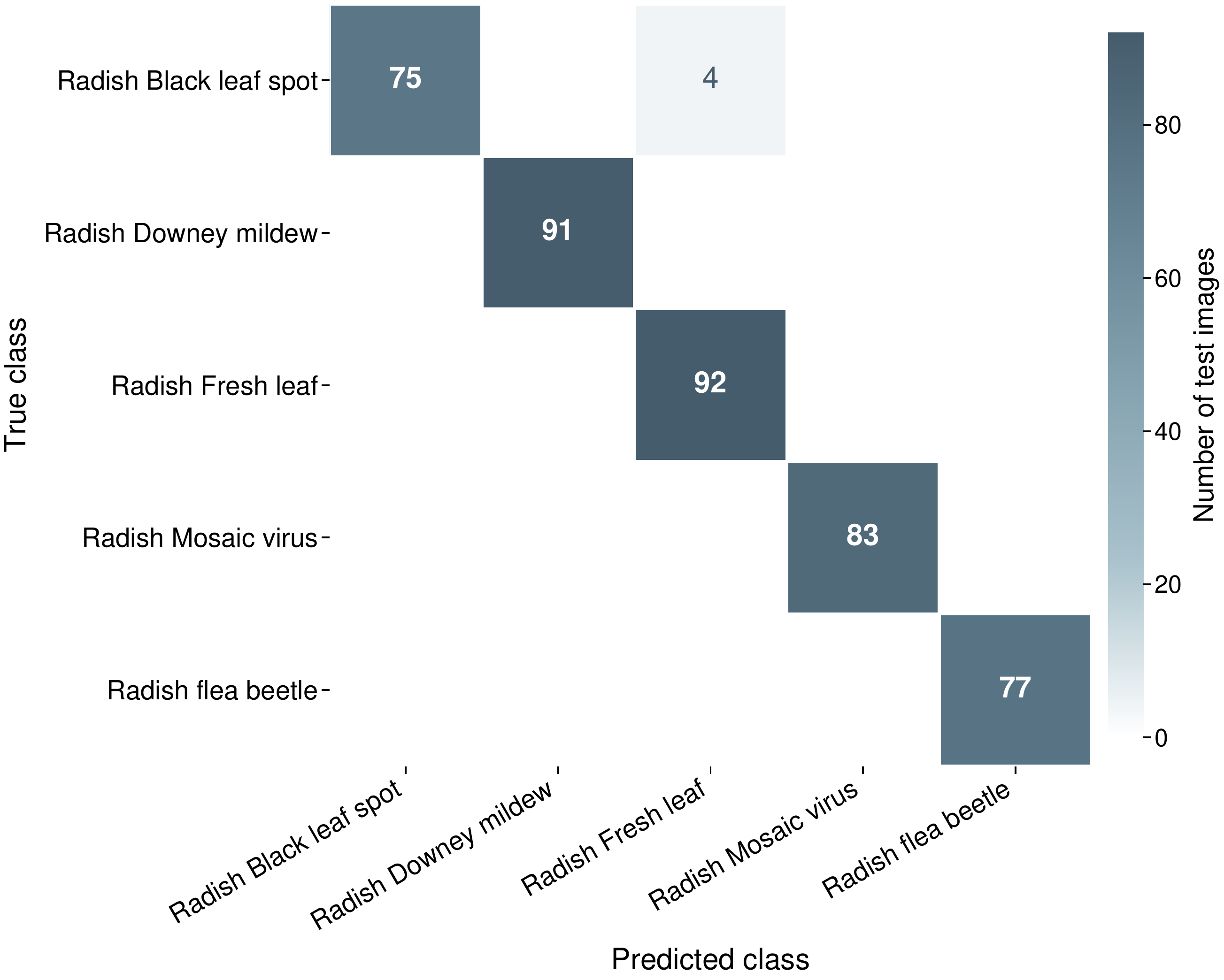}
\caption{Confusion matrix of \nameModel{} on the RadishLeaf-BD test split. Four of 422 images are misclassified.}
\label{fig:radish_cm}
\end{figure}

Figure~\ref{fig:vegnet} decomposes the VegNet-BD result. Twelve of the 21 classes are classified without error, and the aggregate is limited by four bitter gourd classes: Fusarium wilt at 87.18\% F1, mosaic virus at 88.76\%, downy mildew at 98.84\% and the healthy leaf at 98.80\%. Fusarium wilt and mosaic virus both present as interveinal chlorosis on bitter gourd foliage and are separated in the field by root inspection rather than by leaf appearance, and the residual error concentrates where the visual evidence is genuinely ambiguous. Every class of the remaining five crops exceeds 98.9\% F1. Figure~\ref{fig:radish_cm} shows the RadishLeaf-BD confusion matrix, in which the only error is four Radish Black Leaf Spot images assigned to Radish Fresh Leaf out of 422 test images.

A separate-retraining protocol establishes that the architecture transfers to each collection in isolation but leaves open whether it scales to a single heterogeneous label space that spans all three crops at once. To settle this, the three datasets were merged into one corpus and the architecture retrained on it with, again, only the output width altered. The two classes common to \nameDataset{} and RadishLeaf-BD under matching definitions, the healthy radish leaf and the radish flea beetle damage, were unified under one label each, so that the union of the 12, 21 and 5 class inventories collapses to a 36-class taxonomy over 28,019 images. On the resulting 4,223-image stratified test split \nameModel{} attains 99.19\% accuracy, 99.21\% weighted precision and 99.19\% weighted F1 at 292,132 parameters (Table~\ref{tab:cross}, Figure~\ref{fig:cross}), an increase of only 1,560 parameters, or 0.5\%, over the \nameDataset{} configuration and attributable entirely to the wider softmax. Sixteen of the 36 classes are classified without error and 34 of the 4,223 test images are misclassified. The merged accuracy exceeds the model's own standalone figures on VegNet-BD and RadishLeaf-BD and falls 0.33 percentage points short of the standalone \nameDataset{} result. That gap does not originate in the \nameDataset{} classes, which remain nearly intact at 99.46\% accuracy within the merged model against 99.52\% standalone; it is carried by a single pre-existing ambiguous pair from VegNet-BD, bitter gourd Fusarium wilt at 89.51\% F1 with 12 of its 76 test images assigned to bitter gourd mosaic virus, the same confusion reported for that pair in the standalone VegNet-BD experiment where their F1 was 87.18\% and 88.76\% respectively. The architecture therefore scales to a substantially larger, heterogeneous multi-crop label space without any structural change, at a negligible parameter cost, and the accuracy it concedes is confined to genuinely ambiguous classes rather than spread across the taxonomy.

\subsection{Robustness to Input Corruption}
\label{sec:corruption}
\begin{figure}[t]
\centering
\includegraphics[width=0.5\linewidth]{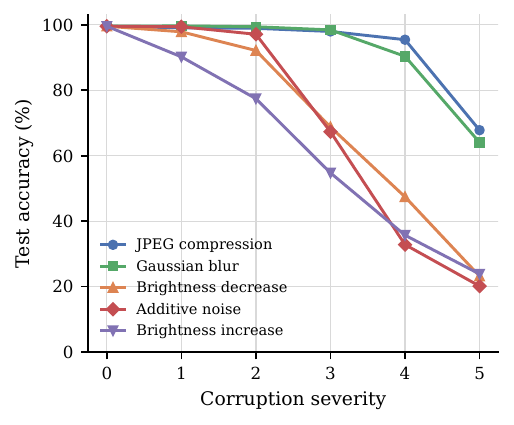}
\caption{Test accuracy of \nameModel{} against corruption severity for five synthetic corruptions applied to the \nameDataset{} test split; severity 0 is the uncorrupted image.}
\label{fig:corruption}
\end{figure}

Field deployment exposes the model to capture conditions absent from the curated test split. To characterize this without new data collection, we applied five synthetic corruptions, Gaussian blur, additive noise, JPEG compression, and brightness decrease and increase, at five increasing severities to the test split and re-evaluated the frozen model; severity~0 is the uncorrupted image at 99.57\% accuracy. Table~\ref{tab:corruption} and Figure~\ref{fig:corruption} report the sweep.For each corruption, the mean drop is computed as the clean accuracy minus the mean accuracy over severities 1 through 5, and Table~\ref{tab:corruption} orders the five corruptions from the most to the least robust by this measure.

The model is most robust to the two corruptions closest to its acquisition pipeline. JPEG compression and Gaussian blur cost only 7.72 and 9.18 percentage points of mean accuracy across the five severities and hold above 95\% and 90\% respectively through severity~4, collapsing only at the most extreme severity~5 (67.81\% and 64.06\%); this is expected, since the corpus is stored as JPEG and moderate blur leaves lesion texture intact. Robustness to additive noise and to illumination shift is markedly weaker. Additive noise holds to severity~2 (97.11\%) but then falls off sharply, to 67.33\% at severity~3 and 20.16\% at severity~5. Increasing brightness is the single most damaging corruption, with additive noise and decreasing brightness following: increasing brightness costs accuracy from the first severity (90.21\%) and reaches 23.85\% at severity~5 for a 43.18-point mean drop, and decreasing brightness follows a similar trajectory to 23.21\%. This asymmetric profile is the signature of the controlled-illumination acquisition protocol of Section~\ref{sec:dataset}: the model never saw strong sensor noise or large exposure changes during training, so these are the genuine out-of-distribution shifts, whereas compression and mild blur preserve the chromatic and textural cues the decision rests on and are tolerated. Test-time augmentation over brightness and noise, or a modest amount of the same augmentation at training time, is the natural mitigation and is deferred to the field-condition study.

\begin{table}[t]
\centering
\caption{Robustness of \nameModel{} to five synthetic input corruptions on the BD-PlantDX test split (Section~\ref{sec:corruption}). Severity 0 is the uncorrupted image at 99.57\% accuracy.}
\label{tab:corruption}
\small
\setlength{\tabcolsep}{4pt}
\begin{tabular}{lrrr}
\toprule
Corruption & Acc.\ at & Mean acc. & Mean drop \\
 & sev.\ 5 (\%) & sev.\ 1--5 (\%) & (pp) \\
\midrule
JPEG compression    & 67.81 & 91.85 & \phantom{0}7.72 \\
Gaussian blur       & 64.06 & 90.40 & \phantom{0}9.18 \\
Brightness decrease & 23.21 & 65.88 & 33.69 \\
Additive noise      & 20.16 & 63.37 & 36.20 \\
Brightness increase & 23.85 & 56.40 & 43.18 \\
\bottomrule
\end{tabular}
\end{table}

\subsection{Interpretability}

Accuracy on a held-out split does not reveal which pixels a model uses, and because \nameDataset{} is acquired against a uniform background a recognizer that keyed on the background could score well and fail on field imagery, a failure mode that accuracy alone cannot detect (RQ5). We therefore combine qualitative attribution maps with a quantitative faithfulness analysis and a sanity check, organized below as method, qualitative attribution, its stage-wise refinement, a failure analysis, a faithfulness study and a model-sensitivity sanity check.

\par\textbf{(i) Method.} We apply Grad-CAM~\citep{selvaraju2017grad} and its generalization Grad-CAM++~\citep{chattopadhay2018gradcampp} to the last convolutional tensor, whose $k$-th activation map is $\mathbf{A}_{k}$. Grad-CAM weights each map by the spatially averaged gradient of the class score and forms the localization map as the rectified weighted sum of Eq.~\eqref{eq:gradcam}, which is min--max normalized and bilinearly upsampled to the $224 \times 224$ input grid. Grad-CAM++ replaces the uniform spatial average of that channel weight with a positive, pixel-wise weighting derived from higher-order derivatives of the class score, which sharpens localization when several disjoint regions support the same class. The two methods therefore probe the same evidence at different spatial granularity, and the faithfulness study of part~(v) determines which is the more representative of how this backbone aggregates it.

\par
\begin{figure}[t]
\centering
\includegraphics[width=0.6\linewidth]{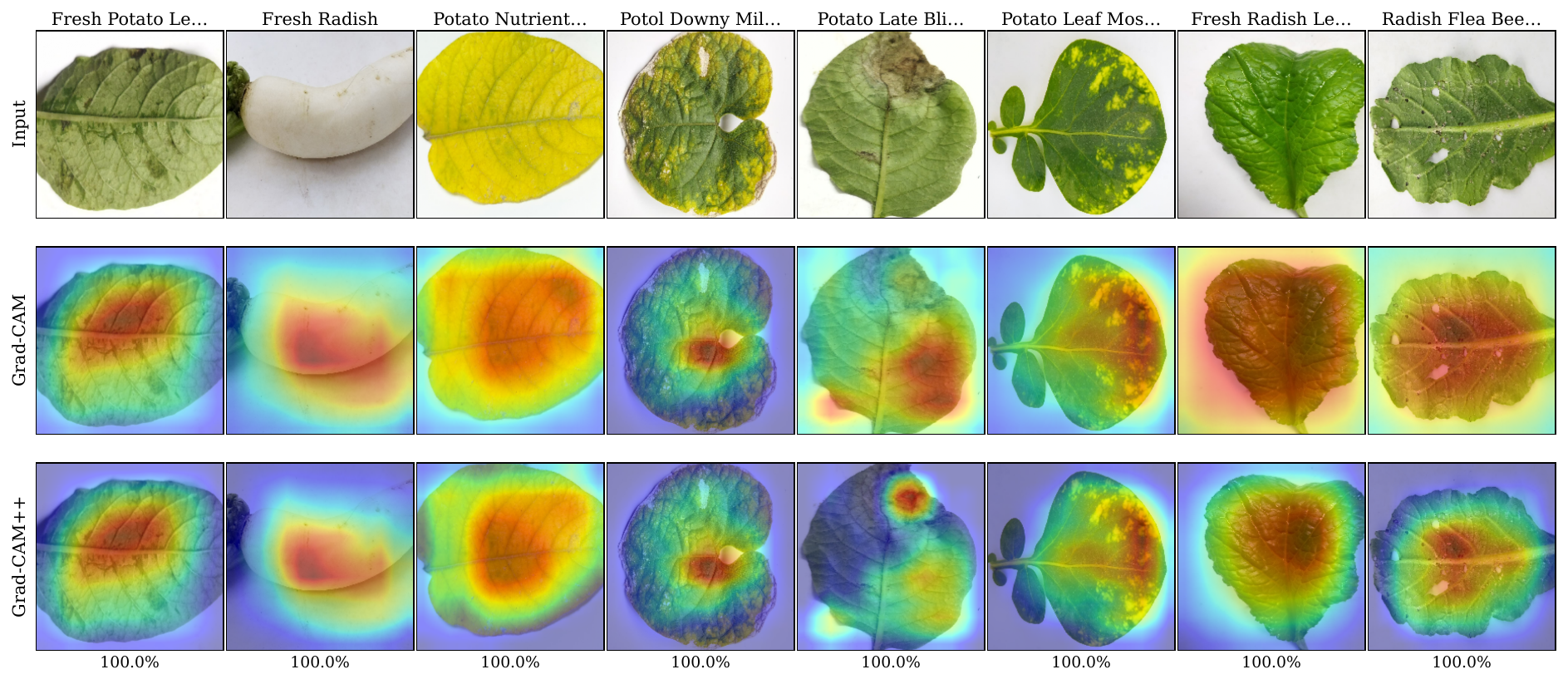}
\caption{Grad-CAM and Grad-CAM++ localization maps of \nameModel{} for eight representative \nameDataset{} classes, with predicted-class probability.}
\label{fig:gradcam}
\end{figure}

\textbf{(ii) Qualitative attribution.} Figure~\ref{fig:gradcam} shows Grad-CAM and Grad-CAM++ maps for eight representative classes, each predicted with essentially full confidence. In every case the attribution peak falls on the leaf lamina, concentrating on the chlorotic, necrotic or mildew-covered tissue that carries the class signal and falling away over the uniform acquisition background. This is direct evidence that the model has not latched onto the controlled background as a class-correlated shortcut, the failure mode to which laboratory-acquired collections are most exposed, and it helps explain the cross-dataset transfer of Section~\ref{sec:results}. Grad-CAM++ produces the more compact maps and Grad-CAM the broader ones; the faithfulness analysis of part~(v) shows the broader maps to be the more representative of how this backbone aggregates evidence.

\par
\begin{figure}[t]
\centering
\includegraphics[width=0.4\linewidth]{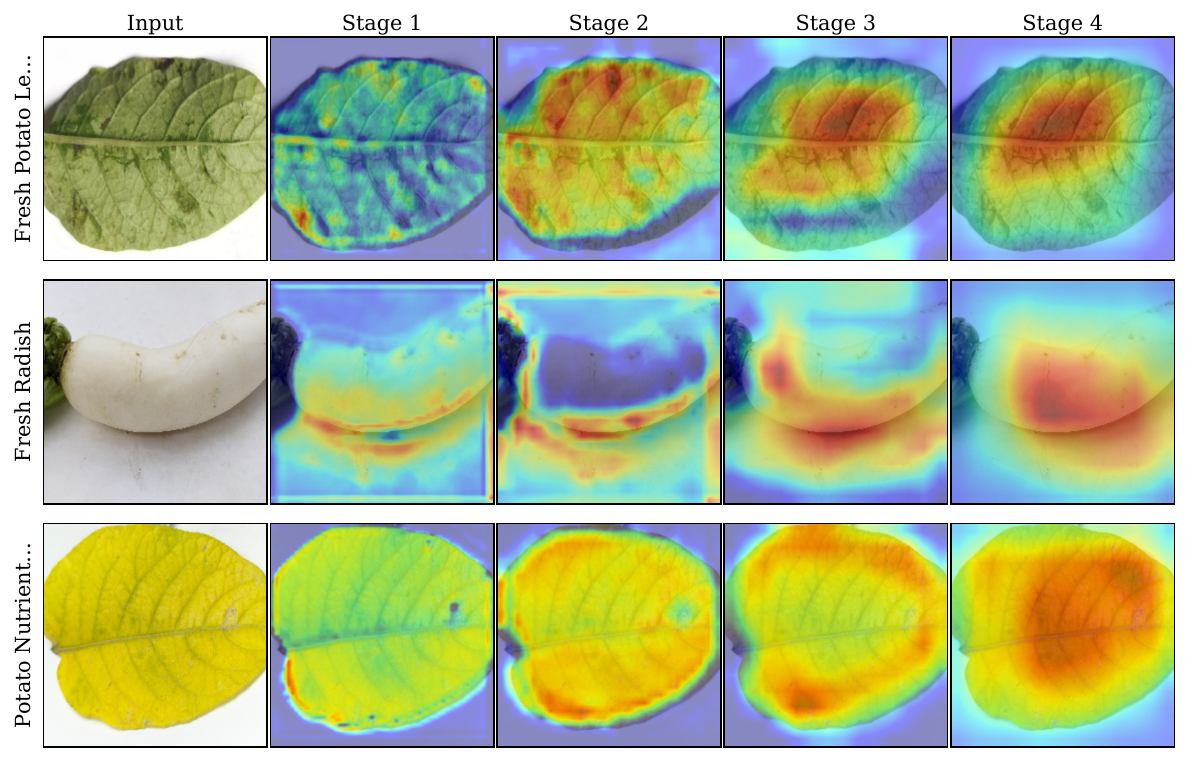}
\caption{Stage-wise evolution of the Grad-CAM map for three classes.}
\label{fig:gradcam_stagewise}
\end{figure}

\textbf{(iii) Where the attribution refines.} Figure~\ref{fig:gradcam_stagewise} traces the map through the four stages of the network. The first stage responds broadly along leaf contours and venation with little class preference; the second begins to separate the affected lamina from the background; and the last two stages consolidate the response onto the symptomatic region. The progression from a generic contour response to a single consolidated region indicates that the class-discriminative representation is built in the deeper stages rather than read off low-level edges.

\par
\begin{figure}[t]
\centering
\includegraphics[width=0.6\linewidth]{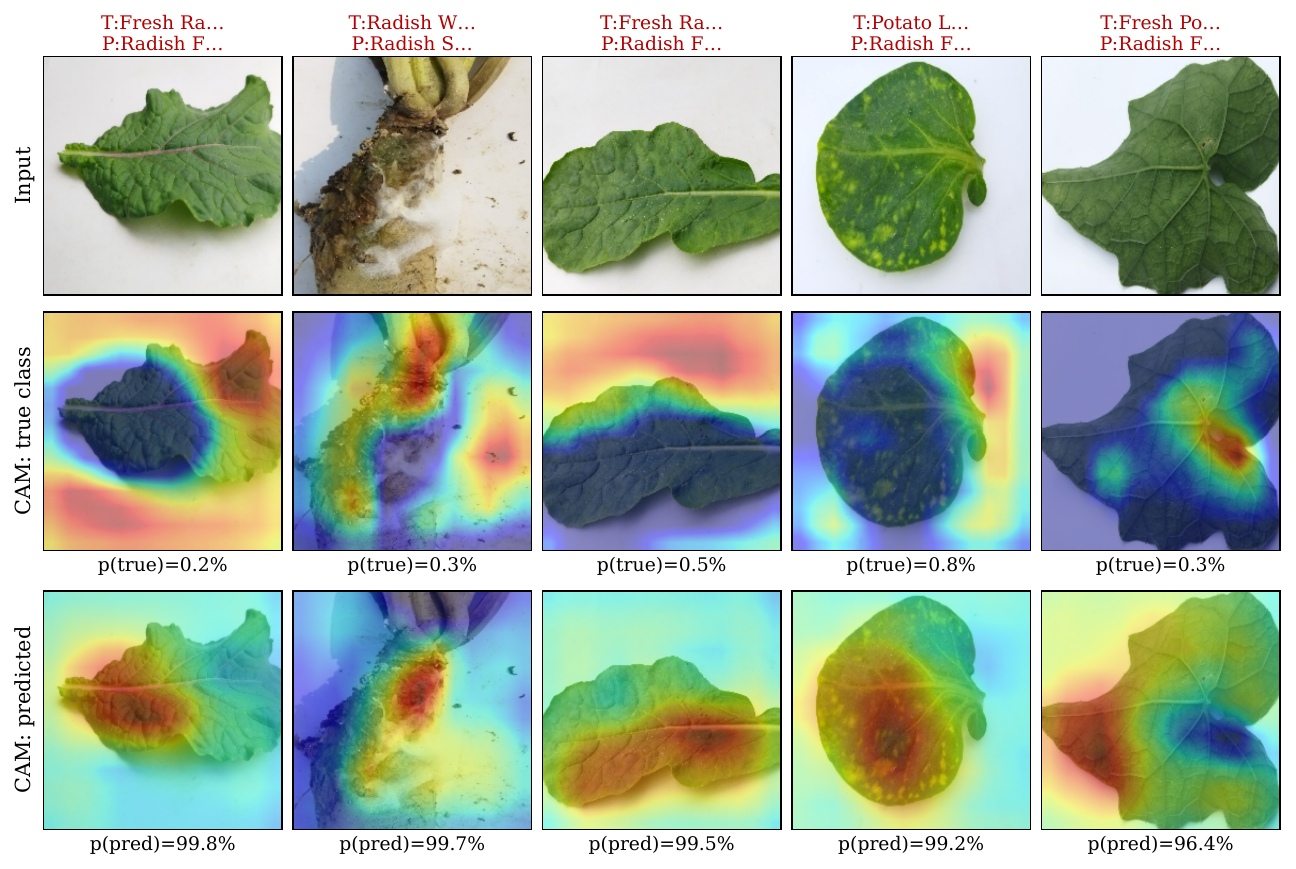}
\caption{Failure analysis of five misclassified test images: Grad-CAM maps for the ground-truth and predicted class, with class probabilities.}
\label{fig:gradcam_failure}
\end{figure}

\textbf{(iv) Failure analysis.} Figure~\ref{fig:gradcam_failure} renders, for five misclassified test images, the Grad-CAM map for the true class alongside that for the predicted class. In all five the two maps fall on the same leaf, so the model localizes correctly and fails at discrimination rather than at attention. The errors are dominated by a single directional confusion: of the nine misclassified test images, seven are assigned to Radish Flea Beetle Damage, four of them from Radish Leaf Healthy and the remainder from Potato Mosaic and Gourd Healthy (Figure~\ref{fig:confusion}), because flea-beetle feeding manifests as sparse millimetre-scale pinholes that a healthy leaf carrying incidental mechanical damage closely mimics at $224 \times 224$ resolution. The model is confidently rather than marginally wrong: mean probability 0.89 on the predicted class against 0.09 on the true class, with five of the nine errors above 0.9 and four above 0.99 on the wrong class. The residual error is therefore a genuine visual near-degeneracy between the flea-beetle and healthy-leaf signatures rather than a background-attention failure, and higher acquisition resolution over the affected region, not stronger attention, is the appropriate remedy.

\par
\begin{figure}[t]
\centering
\includegraphics[width=0.6\linewidth]{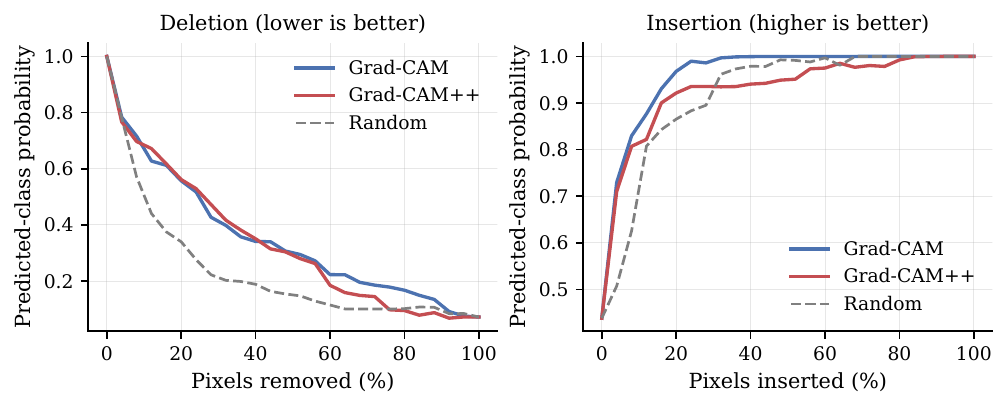}
\caption{Deletion and insertion faithfulness curves for Grad-CAM, Grad-CAM++ and a random control, averaged over 200 test images.}
\label{fig:faithfulness}
\end{figure}

\textbf{(v) Faithfulness.} A plausible-looking map is not itself evidence of faithfulness. We therefore compute the deletion and insertion measures of \citet{petsiuk2018rise} over 200 test images, progressively removing pixels from or restoring them to a blurred baseline in order of decreasing attribution and recording the area under the resulting predicted-probability curve, together with the average confidence drop and increase-in-confidence rate of \citet{chattopadhay2018gradcampp} obtained by masking the input with the normalized map (Table~\ref{tab:faithfulness}, Figure~\ref{fig:faithfulness}). Lower average confidence drop, higher increase-in-confidence, and lower deletion AUC together with higher insertion AUC indicate a more faithful explanation, and no degenerate (uniform or empty) attribution maps were produced across the 200 evaluated images. On insertion, both methods recover the prediction faster than a random-ordering control (insertion AUC 0.932 and 0.925 against 0.902), so the highlighted region is sufficient to support the decision. Grad-CAM is the more faithful of the two attribution methods on the masking metrics, with the lower average confidence drop (41.5\% against 61.6\%) and the higher increase-in-confidence rate, consistent with a compact backbone that aggregates evidence over the whole lamina rather than a few isolated points; all qualitative panels above therefore use Grad-CAM. The deletion measure is the less discriminating here: removing the highest-attribution pixels first does not collapse the prediction faster than random removal (deletion AUC 0.356 and 0.340 against 0.225 for the control), because the leaf fills most of the frame and the evidence is distributed across the lamina, so a compact high-attribution region can be excised while most of the discriminative texture remains. Insertion and the confidence-drop metric, which probe the sufficiency of the highlighted region rather than the effect of a small compact deletion, are the more informative criteria on this leaf-dominated imagery, and both favour the attribution maps over the control.

\begin{table}[t]
\centering
\caption{Faithfulness metrics for attribution maps (200 test images; 50 for random control).}
\label{tab:faithfulness}
\small
\setlength{\tabcolsep}{4pt}
\begin{tabular}{lrrrr}
\toprule
Method & Avg. drop & Increase & Del. AUC & Ins. AUC \\
 & (\%) $\downarrow$ & (\%) $\uparrow$ & $\downarrow$ & $\uparrow$ \\
\midrule
Grad-CAM          & 41.52 & 0.50 & 0.356 & 0.932 \\
Grad-CAM++        & 61.62 & 0.00 & 0.340 & 0.925 \\
Random (control)  & ---   & ---  & 0.225 & 0.902 \\
\bottomrule
\end{tabular}
\end{table}

\par
\begin{figure}[t]
\centering
\includegraphics[width=0.4\columnwidth]{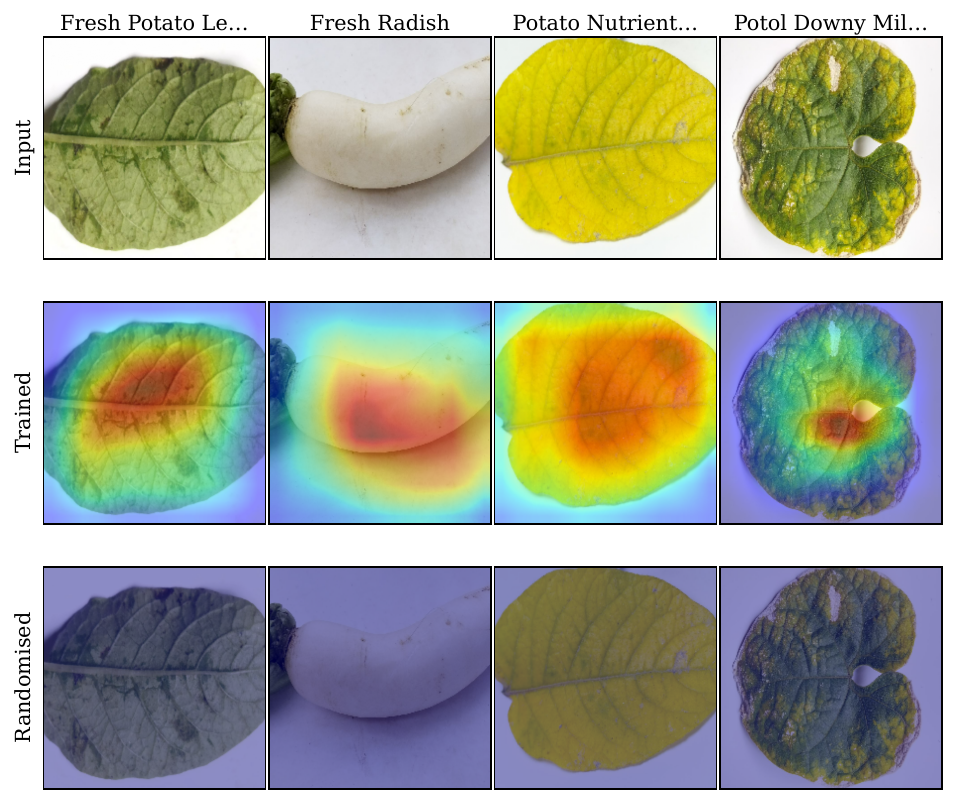}
\caption{Sanity check by cascading weight randomization~\citep{adebayo2018sanity}. input images (top), trained-model Grad-CAM maps (middle), and maps after randomization (bottom).} 
\label{fig:gradcam_sanity}
\end{figure}

\textbf{(vi) Sanity check.} Finally, following \citet{adebayo2018sanity}, we verify that the attribution depends on the learned parameters rather than acting as an edge detector. After cascading randomization of the classifier head and the final stage, the Grad-CAM maps collapse to a near-uniform field with no correspondence to the leaf (Figure~\ref{fig:gradcam_sanity}), unlike the sharply localized maps of the trained model, and they notably do not fall back on the contour structure that dominates the first stage. Localization is therefore a property of the learned weights rather than of the input's edge content, so the explanations satisfy a model-sensitivity criterion that many published saliency analyses fail.

\subsection{Discussion and Limitations}

For \textbf{RQ1}, a convolutional network trained from scratch at 290,572 parameters exceeds every ImageNet-pretrained lightweight backbone evaluated here, reaching 99.52\% against 99.25\% for the strongest lightweight baseline while using 11.2 times fewer parameters, and the five-seed lower bound of 99.47\% remains above all six baselines. Pretraining is therefore not a prerequisite for competitive accuracy at this scale once the architecture matches the structure of the target distribution.

For \textbf{RQ2}, the ablation isolates the swish activation as the largest single accuracy contributor at 0.27 percentage points, and identifies the multi-scale depthwise block and the grouped convolutions as the components that contribute accuracy and parameter economy at once, each of whose removal raises the parameter count by 41.0\% or more while lowering accuracy. The dual-pooling head is accuracy neutral within the ablation protocol at a cost of 16,384 parameters, and augmentation together with weight decay each cost 0.16 percentage points of clean-test accuracy under the current acquisition protocol.

For \textbf{RQ3}, the unmodified architecture reaches 98.71\% on 21 classes of VegNet-BD and 99.05\% on five classes of RadishLeaf-BD, with only the output width altered, which indicates that the design encodes a general inductive bias for foliar lesion discrimination rather than a fit to \nameDataset. The same architecture further sustains 99.19\% accuracy when the three datasets are merged into a single 36-class corpus of 28,019 images at a 0.5\% parameter increase, which confirms that the design scales to a heterogeneous multi-crop label space and not merely to each collection in isolation.

For \textbf{RQ4}, accuracy over five seeds is $99.57 \pm 0.10$\% within a 0.21 percentage point range, while the epoch of best validation performance varies by a factor of 2.2, showing that the reported accuracy is reproducible even though the optimization trajectory is not; a paired analysis over the same five seeds further confirms that the margin over every pretrained baseline is statistically reliable, with \nameModel{} winning all five seeds against each backbone and a Holm-corrected paired $t$-test significant in every case (Section~\ref{sec:significance}).

For \textbf{RQ5}, the attribution maps localize onto lesion-bearing lamina rather than onto background across representative classes and misclassifications alike; the insertion and confidence-drop faithfulness measures confirm that the highlighted region carries the evidence the model uses, and a weight-randomization sanity check confirms that the localization depends on the learned parameters rather than on input edges. The deletion measure does not separate the maps from a random control on this leaf-dominated imagery, and a quantitative agreement against agronomist-drawn lesion masks remains outstanding.

Four limitations bound these conclusions. All \nameDataset{} imagery was acquired under a controlled illumination arrangement against a uniform background, and accuracy under field illumination, cluttered backgrounds and partial occlusion is not yet measured; the synthetic corruption sweep of Table~\ref{tab:corruption} is a proxy that shows tolerance to compression and blur but sensitivity to additive noise and brightness shift, and a genuine field-condition subset remains the outstanding test. The efficiency comparison now covers multiply-accumulate cost, single-precision and quantized model size, desktop CPU inference latency and, in Table~\ref{tab:edge}, on-device latency and throughput on a mid-range Android handset, but sustained-load thermal behaviour and peak activation memory remain open, the latter reported neither on-device nor for the baselines. The paired margin over the baselines is now established by a Holm-corrected $t$-test and a clean five-of-five seed sweep with all six confidence intervals excluding zero (Section~\ref{sec:significance}); the residual caveat is only that the Wilcoxon signed-rank test is underpowered at five seeds, its exact two-sided $p$-value bounded below by 0.0625, so a larger seed budget would let the rank test corroborate the same margin. Finally, the twelve classes cover the pathologies prevalent in two districts over one cropping season, and coverage of additional agro-ecological zones and seasons remains open.

\section{Conclusion}
\label{sec:conclusion}
We presented \nameModel, a convolutional network of 290,572 trainable parameters that composes grouped bottleneck residual blocks carrying sequential channel and spatial attention, multi-scale depthwise blocks fusing $3 \times 3$ and $5 \times 5$ responses, and a dual-pooling classification head, together with \nameDataset, an expert-validated benchmark of 12,432 field images spanning 12 healthy and diseased classes of radish, potato and pointed gourd collected across the Bogura and Nilphamari districts of Bangladesh. On \nameDataset{} the model attains 99.52\% accuracy and 99.52\% weighted F1, exceeding all six ImageNet-pretrained lightweight backbones evaluated under an identical protocol while using 8.7 to 16.8 times fewer parameters and the fewest multiply-accumulate operations of any model compared, and it exceeds a four-model heavyweight transfer ensemble by 0.16 percentage points at less than one percent of its parameter count. Exported for deployment, the model quantises to a 0.46 MB full-integer network at a 0.22 percentage-point accuracy cost and classifies an image in under ten milliseconds on a single CPU. Accuracy is stable at $99.57 \pm 0.10$\% over five seeds and its margin over every ImageNet-pretrained baseline holds across all five seeds under a paired significance analysis, an eleven-variant ablation shows that the multi-scale depthwise block and the grouped convolutions supply accuracy and parameter economy jointly, the unmodified architecture transfers to two independently collected datasets at 98.71\% and 99.05\% accuracy, and Grad-CAM evidence indicates that predictions rest on lesion-bearing tissue rather than on background cues. The present evaluation rests on imagery acquired under controlled illumination, and the efficiency characterization, while it now spans compute, model size, quantization, and CPU latency, does not yet include measurements on the target hardware itself.

Future work will close these gaps: a field-condition evaluation subset with natural backgrounds and variable illumination will establish whether the accuracy reported here survives deployment conditions, the synthetic corruption sweep of Section~\ref{sec:results} being a first step that shows strong tolerance to compression and blur but sensitivity to additive noise and illumination shift; sustained-load latency and memory characterisation on the target embedded hardware will carry the quantised model from a laboratory efficiency profile to a verified on-device deployment claim; and severity grading annotation will extend the label space from disease presence toward the intervention decision a grower actually faces.

\section*{Declaration of Generative AI and AI-Assisted Technologies in the Manuscript Preparation Process}
\label{sec:llm_usage}
During the preparation of this manuscript, large language models (LLMs) were employed only as supporting tools for improving language and for assisting with code debugging. All AI-assisted material was subsequently examined, verified, and corrected by the authors, who assume full responsibility for the accuracy, originality, and integrity of the work presented here.\par
\paragraph{Originality} The role of the LLMs was strictly editorial. Any passages that were drafted or polished with such assistance were fact-checked and rewritten by the authors to ensure consistency with the actual research findings. Every figure, table, and quantitative outcome, together with all technical descriptions and reported numerical values, was generated solely by the authors without any AI contribution.\par
\paragraph{Transparency} LLMs were additionally applied during code development, mainly for identifying and resolving implementation errors. Even so, the experimental pipeline demanded considerable manual design, incremental correction, and iterative testing, since AI-generated recommendations on their own proved inadequate. Each reported result was produced by executing author-verified code and was checked against expected behaviour prior to inclusion. The underlying methodology, the model architecture, and the AgroVisNet experimental protocol were designed and validated by the authors independently of any LLM.\par
\paragraph{Responsibility} At no point during the writing or debugging process was any sensitive, private, or proprietary information such as dataset images or participant data disclosed to an AI tool, and every interaction adhered to ethical principles concerning data ownership and intellectual property. The use of LLMs remained limited to general writing support and code debugging, and it had no bearing on the scientific contributions or the claims advanced in this paper.

\section*{CRediT authorship contribution statement}

\textbf{Md. Abdullah Mandal:} Conceptualization, Data curation, Formal analysis, Investigation, Methodology, Project administration, Software, Visualization, Validation, Writing - original draft. \textbf{Saad Ahmed:} Formal analysis, Resources, Supervision, Validation, Writing - review \& editing. \textbf{Md. Khalid Syfullah:} Formal analysis, Project administration, Resources, Supervision, Writing - review \& editing.

\section*{Declaration of Competing Interest}

The authors declare that there are no known competing financial interests or personal relationships that could have appeared to influence the work reported in this article.

\section*{Acknowledgments}

The authors gratefully acknowledge Professor Dr.\ Md.\ Harun Ar Rashid,
Department of Horticulture, Bangladesh Agricultural University, Mymensingh,
for his expert validation of the \nameDataset{} dataset. His independent,
per-class review of the radish, potato and pointed gourd images, recorded in
a signed Certificate of Data Integrity, confirmed the accuracy and
consistency of the healthy and diseased class assignments across all twelve
classes. The authors further thank the local agronomists who guided the field
surveys and identified the diseased plots across the Bogura and Nilphamari
districts, and the farmers and landholders who permitted specimen collection
from their fields. Their agronomic expertise and cooperation were essential to
the construction of \nameDataset.The authors received no external funding for this study.
 
\section*{Data and Code Availability}

The BD-PlantDX dataset, the trained AgroVisNet weights, and the figure-generation notebook will be released at \href{https://github.com/Abdullah2104/agrovisnet}{(https://github.com/Abdullah2104/agrovisnet)}.


\bibliographystyle{elsarticle-harv}
\bibliography{references}
\end{document}